\pdfoutput=1

\documentclass[11pt]{article}

\usepackage[final]{acl}

\usepackage{times}
\usepackage{latexsym}

\usepackage[T1]{fontenc}
\usepackage[utf8]{inputenc}

\usepackage{microtype}

\usepackage{inconsolata}

\usepackage{graphicx}
\usepackage{subcaption}
\usepackage{amsmath}
\usepackage{amssymb}
\usepackage{booktabs}
\usepackage{array}
\newcolumntype{P}[1]{>{\centering\arraybackslash}p{#1}}
\usepackage{algorithm,algpseudocode}
\usepackage{siunitx}

\counterwithin{figure}{section}
\counterwithin{table}{section}
\counterwithin{equation}{section}

\title{Uncertainty in Semantic Language Modeling with PIXELS}

\author{\textbf{Stefania Radu}, \textbf{Marco Zullich}, \textbf{Matias Valdenegro-Toro}\\
  Department of Artificial Intelligence, Bernoulli Institute, University of Groningen, The Netherlands \\
  \texttt{stefania.m.radu@gmail.com}, \texttt{m.a.valdenegro.toro@rug.nl}}

\begin{document}
\maketitle
\begin{abstract}
Pixel-based language models aim to solve the vocabulary bottleneck problem in language modeling, but the challenge of uncertainty quantification remains open. The novelty of this work consists of analysing uncertainty and confidence in pixel-based language models across 18 languages and 7 scripts, all part of 3 semantically challenging tasks. This is achieved through several methods such as Monte Carlo Dropout, Transformer Attention, and Ensemble Learning. The results suggest that pixel-based models underestimate uncertainty when reconstructing patches. The uncertainty is also influenced by the script, with Latin languages displaying lower uncertainty. The findings on ensemble learning show better performance when applying hyperparameter tuning during the named entity recognition and question-answering tasks across 16 languages.
\end{abstract}

\section{Introduction}

After the release of ChatGPT in $2022$, the number of papers published every day on the topic of Large Language Models (LLMs) has increased more than 20-fold \citep{zhao2023survey}. The number of parameters in these models jumped from $340$ millions in implementations such as BERT \citep{devlin2018bert} to billions of parameters in models like GPT-3 \citep{brown2020language} or LLaMA \citep{touvron2023llama}. Despite their obvious popularity, one of the central limitations of LLMs remains their uncertainty and lack of trustworthiness \cite{huang2024trustllmtrustworthinesslargelanguage}. As these models are being applied more and more to high-stakes scenarios, such as medicine \cite{busch2025current} or security \citep{ gawlikowski2023survey}, it is critical that their predictions can be trusted. Generally, the research on the explainability and interpretability of LLMs is focused on traditional tokenizer-based methods, that split text into smaller units. They produce overconfident responses even when the predictions are likely incorrect \citep{xiong2023can}. 

\begin{figure}[t]
    \centering
    \includegraphics[width=0.8\linewidth]{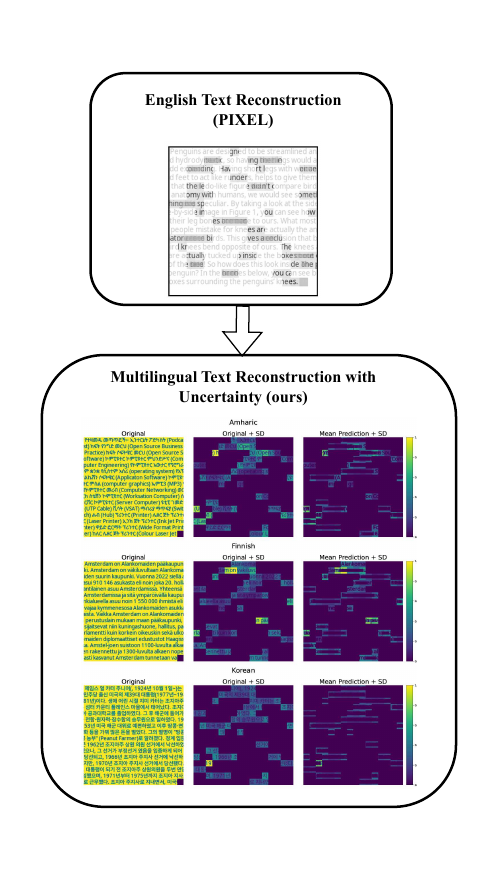}
    \caption{Example of text reconstruction using the PIXEL model from \citet{rust2022language}, and text reconstruction with uncertainty for different languages.}
\label{fig:intro}
\end{figure}

For semantic NLP tasks such as extractive question answering (QA), it is common to use models that predict the start and end tokens of an answer span and provide confidence scores based on the softmax probabilities of these predictions \citep{devlin2018bert, lan2019albert}. However, this approach offers no measure to quantify the uncertainty of the prediction. Several works have been proposed in the past years to solve this problem \citep{xiao2022uncertainty, lin2023generating}. Common solutions include incorporating uncertainty directly into the model using Bayesian Neural Networks (BNN) \citep{blundell2015weight} or post-hoc methods such as Monte Carlo Dropout \citep{gal2016dropout}, Temperature Scaling \citep{guo2017calibration} and Ensemble Learning \citep{lakshminarayanan2017simple}. However, these approaches have not been studied in the context of more recent pixel-based models that use visual representations of words, as opposed to text representations.

The \textit{Pixel based Encoder of Language} or PIXEL proposed by \cite{rust2022language} aims to transform language modeling into a visual recognition task with the help of small and square clusters of pixels, called \textit{patches}. PIXEL does not rely on a predefined vocabulary and it is trained to reconstruct missing patches of text, by following a Vision Transformer -- Masked Autoencoder (ViT-MAE) architecture. The Vision Transformer (ViT) uses linear embeddings of fixed-sized patches of pixels which are encoded using a transformer. In the context of computer vision, masked image encoding works similarly to masked language modeling (MLM), by masking regions of an image and then learning to reconstruct the whole image. 

PIXEL was pretrained on rendered versions of the Wikipedia and BookCorpus datasets and it is evaluated on 32 topologically diverse languages, across 14 scripts. Supporting multiple languages requires a larger vocabulary to cover diverse
linguistic features and scripts, which is often impractical within the constraints of a
fixed vocabulary size. \citet{wu2019beto} noted that multilingual models struggle
with resource allocation across languages, leading to suboptimal performance in less
represented languages, during tasks like named entity recognition, part-of-speech tagging, and dependency parsing. Furthermore, imbalanced vocabulary representation
can exacerbate biases, resulting in unfair treatment of certain languages  \citep{wan2021fairness}.
The trade-off in vocabulary allocation means that models either inadequately repre
sent some languages or become too large in size and computational requirements.

The main aim is to study uncertainty in pixel-based language models focusing on semantic tasks. Given the challenging nature of semantic processing and the fewer studies dedicated to it, this research will center on finetuning models to solve tasks like named entity recognition, sequence classification, and question answering. Solving the vocabulary bottleneck of traditional language models which rely on a close vocabulary can be achieved by using pixel-based models which do not require a fixed vocabulary. Finally, to tackle the uncertainty problem, this work will make use of existing techniques for quantifying uncertainty, and apply them to pixel-based models, which also represent the biggest novelty of this study. This includes uncertainty quantification at the pixel level using Monte Carlo methods (Figure \ref{fig:intro}), ensemble learning applied to models finetuned on three semantic tasks across 19 languages, but also an analysis of the attention mechanism.

\section{State of the Art}

The first study to use visual features of text in order to create embeddings was applied to Chinese and used linearizing bitmaps of characters or words \citep{aldon2016neural}. By using shared character components from Chinese or Korean, it becomes easier to generalize to new and less frequent characters. Different studies \cite{dai2017glyph, sun2018super, salesky2021robust} used rendering techniques to obtain images of text. In this context, text rendering involves converting character codes into glyph indices, which are then used to generate the corresponding glyph images, while applying various styles, fonts, sizes, and colors. A glyph often contains one character only, but it can also represent accents or multiple characters in languages where ligatures are common, like Arabic. \citet{dai2017glyph} used text rendering in Chinese, Japanese, and Korean, and extracted visual features from a Convolutional Neural Network (CNN) to perform text classification. Similarly, \citet{sun2018super} applied convolutions to squared rendered images to perform sentiment analysis in Chinese and English. 

In the context of machine translation, \citet{salesky2021robust} suggested a very robust approach based on a variation of the ViT. The training data is rendered into gray-scale images using the Pygame backend and a slicing window is applied to create patches, which act as tokens. Then, a 2D convolutional block followed by linear projection is used to create embeddings, which serve as input for the transformer encoder. The translation happens directly from pixel representations, without any word preprocessing. After training on seven language pairs, the approach matches the performance of traditional language models, with additional advantages. It is more robust to character permutations or substitutions, and it does not rely on text preprocessing steps, such as tokenization or segmentation.

As of to date, systematic investigations into the uncertainty and calibration of pixel-based language models remain limited. \citet{rust2022language} showed that PIXEL is robust when it comes to character-level perturbations and code-switching. In this analysis, relevancy heatmaps were used to depict visual explanations of correct predictions, and there is evidence to suggest that these outputs are interpretable when identifying contradictions and entailment relationships. However, during semantic tasks like named entity recognition, sequence classification, and question answering, PIXEL is struggling to retain semantic knowledge and transfer it across scripts. Reasons for this might include a lack of multilingual pretraining, as well as a limited ability to capture contextual information due to the use of unigram patch embeddings. While raw performance is desirable, it is crucial to have models that are reliable and explainable.

\begin{figure*}[t]
    \centering
    \begin{subfigure}[b]{0.27\textwidth}
        \centering
        \fbox{\includegraphics[width=\linewidth]{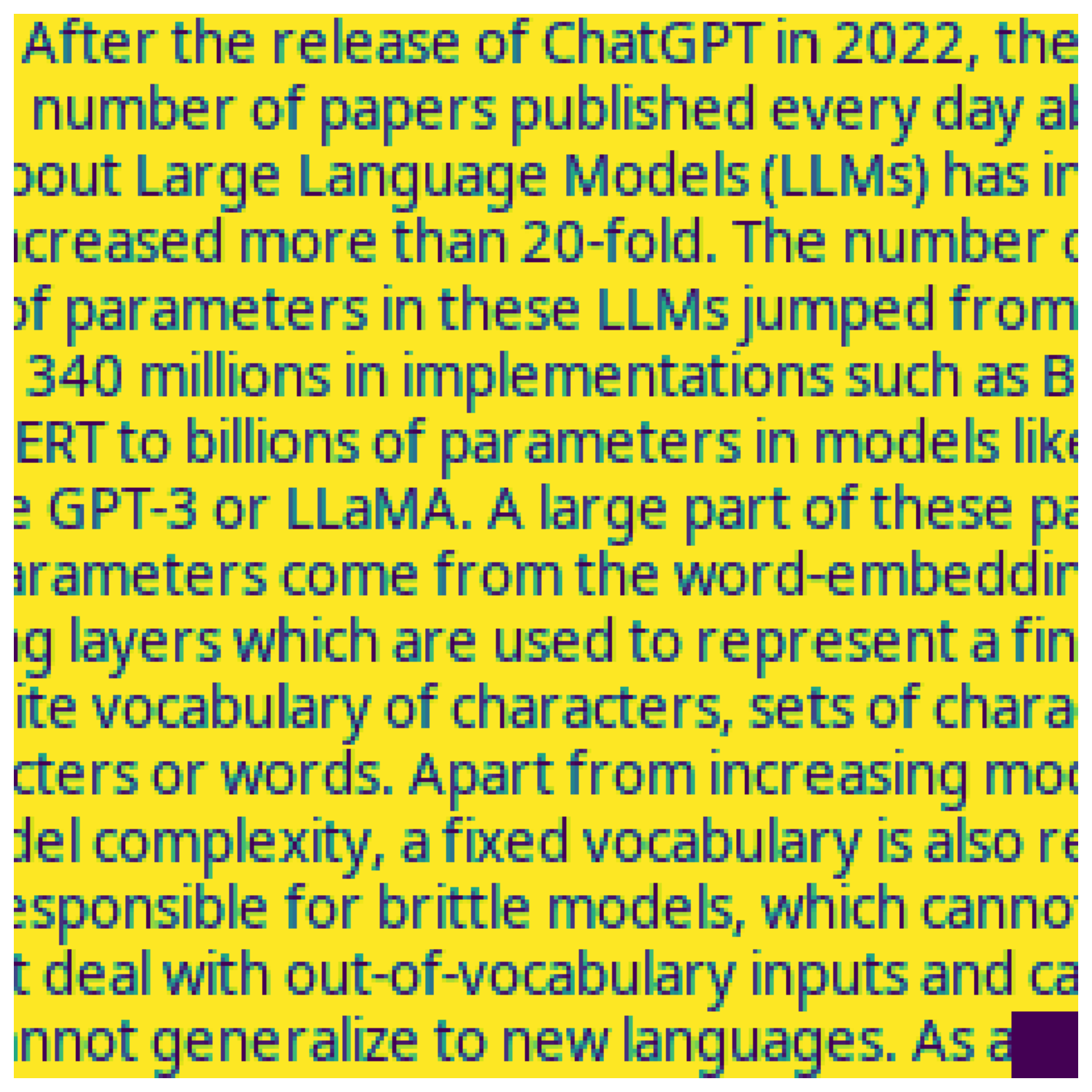}}
        \caption{Original rendered text using the PyGame renderer.}
        \label{fig:image1}
    \end{subfigure}
    \hfill
    \begin{subfigure}[b]{0.27\textwidth}
        \centering
        \fbox{\includegraphics[width=\linewidth]{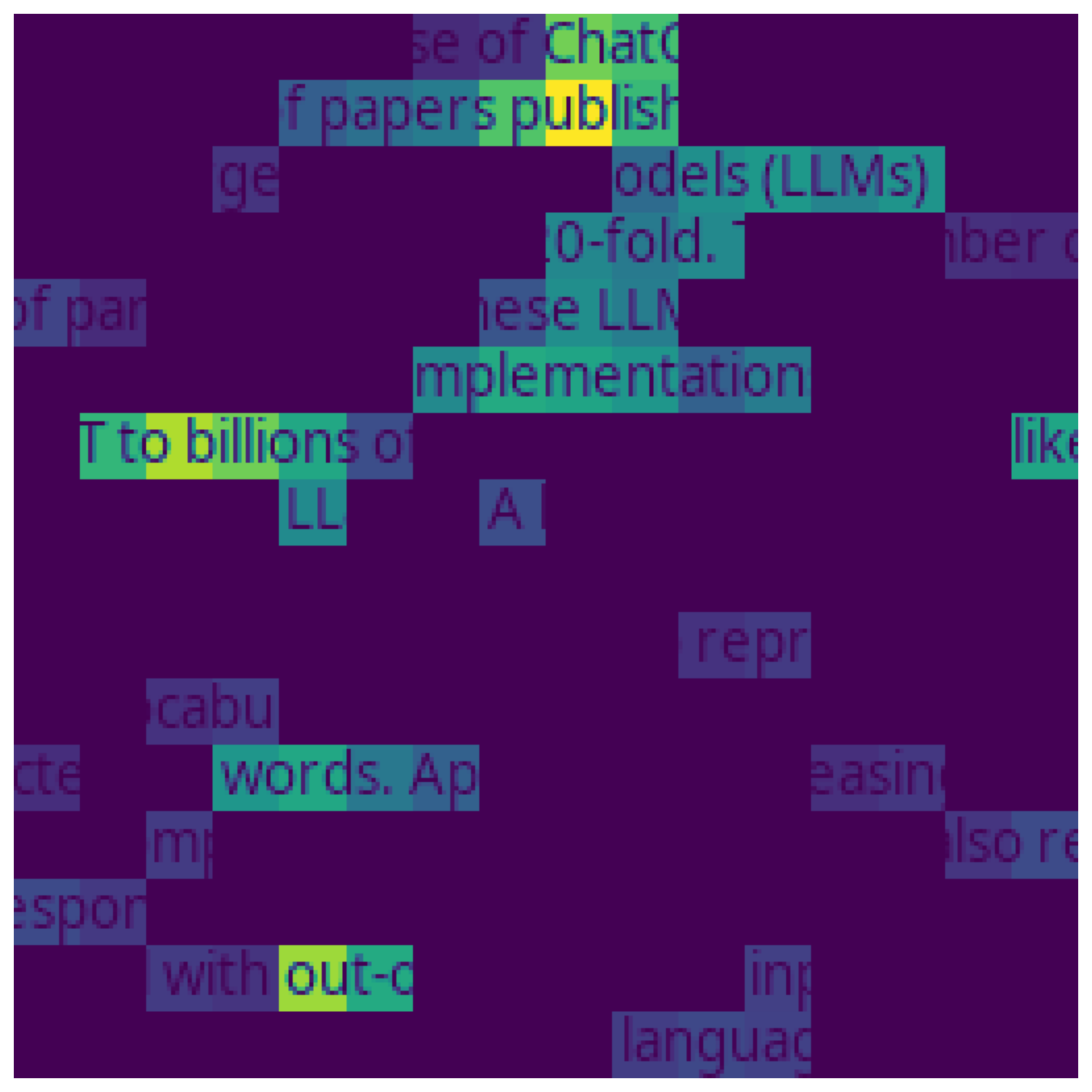}}
        \caption{Original image with uncertainty.}
        \label{fig:image2}
    \end{subfigure}
    \hfill
    \begin{subfigure}[b]{0.27\textwidth}
        \centering
        \fbox{\includegraphics[width=\linewidth]{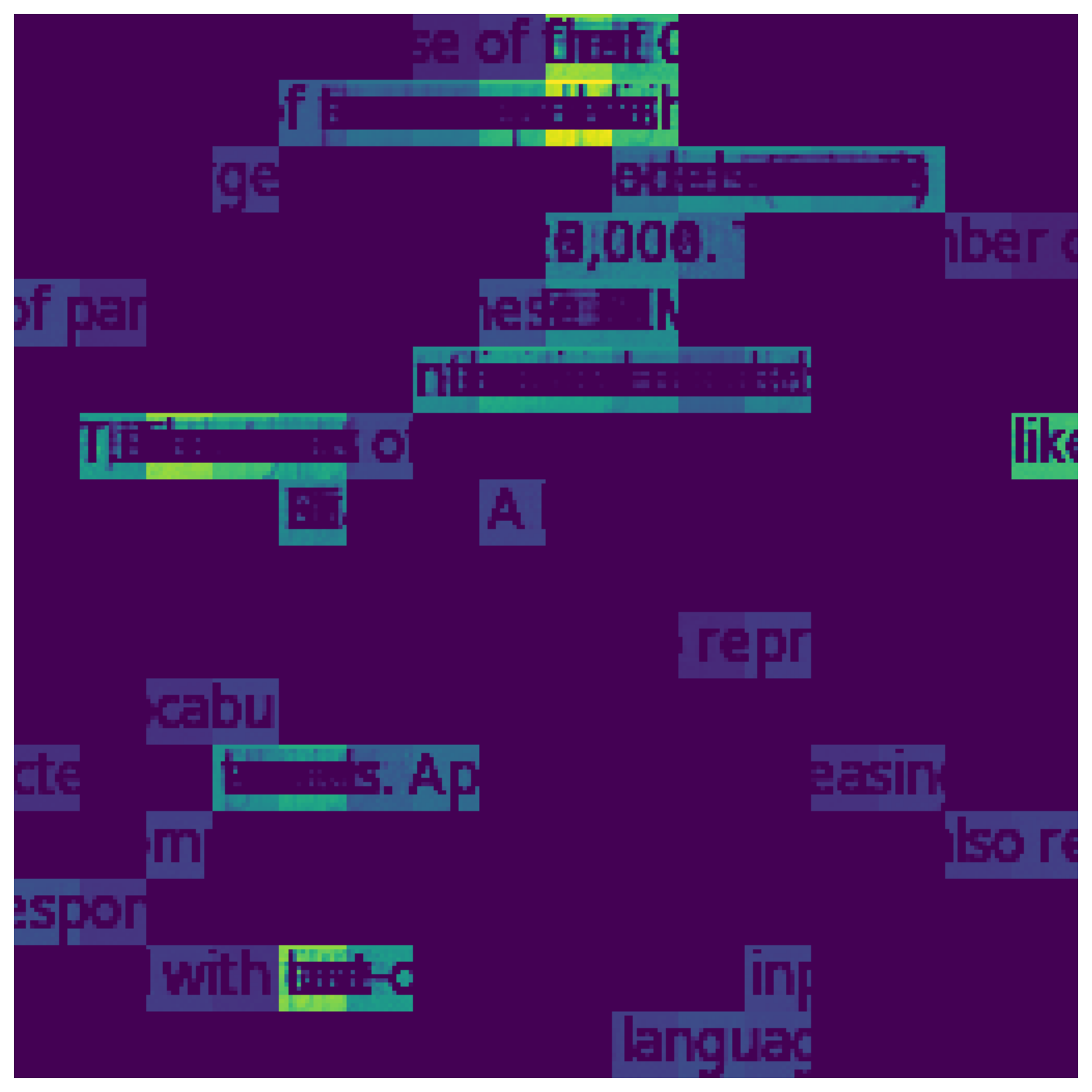}}
        \caption{Reconstructed text with uncertainty.}
        \label{fig:image3}
    \end{subfigure}
    \hfill
    \begin{subfigure}[b]{0.04\textwidth}
        \centering
        \includegraphics[width=\linewidth]{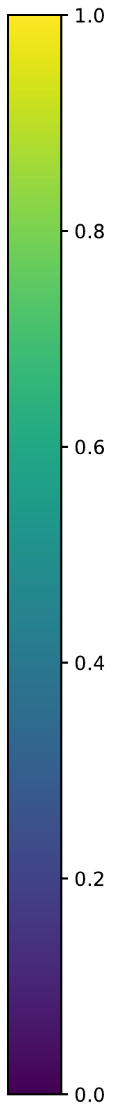}
        \caption*{}  
    \end{subfigure}

    \caption{Example of uncertainty quantification at the patch level for an image 
             containing text from the introduction of this paper. Brighter colors indicate 
             more uncertainty.}
    \label{fig:MC-uncertainty-first}
\end{figure*}

\section{Methods}

\subsection{Data}

\textbf{MasakhaNER 1.0}
\label{sec-data-MasakhaNER} 
MasakhaNER 1.0 \citep{adelani2021masakhaner} is a Named Entity Recognition (NER) benchmark, which includes data from 10 African Languages obtained from local news sources (Amharic, Hausa, Igbo, Kinyarwanda, Luganda, Luo, Nigerian-Pidgin, Swahili, Wolof and Yorùbá), as well as the ConLL-2003 English dataset. The task involves classifying named entities into nine pre-defined categories. The MasakhaNER dataset contains labeled entities for each language. 

\textbf{GLUE}
\label{sec-data-glue}
The Sequence Classification (SC) task relies on the The General Language Understanding Evaluation (GLUE) benchmark \citep{wang2018glue}. It involves nine sentence-level understanding tasks (CoLA, SST-2, MRPC, QQP, STS-B MNLI-M/MM, QNLI, RTE, WNLI) in English, across three categories: single-sentence tasks, similarity and paraphrase tasks, and inference tasks.

\textbf{TyDiQA-GoldP}
\label{sec-data-TyDiQA}
To assess the ability of the model to perform Question Answering (QA), the TyDiQA-GoldP dataset was selected \citep{clark2020tydi}. It contains nine typologically diverse languages (English, Arabic, Bengali, Finnish, Indonesian, Korean, Russian, Swahili, Telugu). The dataset contains questions written by native speakers, passages with relevant information, and answers provided as short spans of text within the passage. Unlike the primary task, the Gold Passage task focuses more on locating the exact answer within a given context. 

\subsection{Model Architecture}

PIXEL processes text as images that are rendered using the PyGame\footnote{\url{https://www.pygame.org/}} renderer to accommodate multiple scripts. Each rendered image is converted into a sequence of patches, resulting in $529$ non-overlapping patches, with a size of $16 * 16$ pixels. A ViT-based encoder encodes visible patches and the CLS tokens through patch, positional, and CLS embeddings. During pretraining, the system applies random masking to $25\%$ of the patches and employs a decoder to reconstruct the masked regions through a regression-like method. The decoder is then finetuned on downstream tasks by replacing the reconstruction objective with task-specific heads.

The English PIXEL which serves as a base for the experiments described in the next section is pretrained on a rendered version of English Wikipedia and BookCorpus \citep{zhu2015aligning}. For more details about the PIXEL pretraining routine, refer to the implementation\footnote{\url{https://github.com/xplip/pixel}} of \citet{rust2022language}.

\subsection{Uncertainty Quantification}

\textbf{Monte Carlo Uncertainty}
The first method used to quantify epistemic uncertainty at the patch level is Monte Carlo (MC) Dropout. The input is a rendered \(\texttt{image} \in \mathbb{R}^{16 \times 16 \times 3}\) with a sequence length of 256 pixels, and the goal is to obtain an uncertainty map $U \in \mathbb{R}^{16 \times 16 \times 3}$, containing the uncertainty for each patch. For this, the model is used in 100 forward passes to compute a series of predictions $P$, which contain per-pixel logits. Then, the mean prediction is created by averaging these logits, resulting in the reconstructed text. A standard deviation (SD) image is obtained by computing the SDs of the predictions for each pixel. Since each patch has a dimension of $16 \times 16$ pixels, the per-patch uncertainty is defined by averaging the predictions of all SD values inside a patch, and each pixel inside the patch is assigned that value. Finally, the uncertainty map $U$ is a collection of patches representing the overall uncertainty of its pixels. For visualization purposes, the uncertainty map is overlaid on top of the original image, as well as on the reconstructed text. An overview of this routine is presented in Algorithm \ref{alg:MC-uncertainty} of Appendix \ref{sec:appendix-experiments}.

An overall mean uncertainty value ($\bar{\sigma}$) is also computed to measure uncertainty at the image level (Equation \ref{eq:mean-uncertainty}), where $H$ and $W$ refer to the height and width of the image.

\begin{equation}
\bar{\sigma} = \frac{1}{H \times W} \sum_{h=1}^{H} \sum_{w=1}^{W} \sigma(h, w)
\label{eq:mean-uncertainty}
\end{equation} 

Additionally, we compute two loss functions during the MC inference: the normalized MSE loss (Equation \ref{eq:mse-loss})  used during pretraining and the normalized Gaussian Negative Log-Likelihood (GNLL) loss (Equation \ref{eq:gnll-loss}), where $eps=1e-6$ is a clamp value used for stability. Unlike the MSE, the GNLL loss accounts for epistemic uncertainty, by incorporating the variance of the predicted distribution.

\begin{equation}
    \text{MSE} = \frac{1}{H \times W }  (\text{pred} - \text{img})^2
    \label{eq:mse-loss}
\end{equation}

\begin{equation}
    \text{GNLL} = \log (\max (\text{var}, \text{eps})) + \frac{(\text{pred} - \text{img})^2}{\max (\text{var}, \text{eps})} 
    \label{eq:gnll-loss}
\end{equation}

We study uncertainty across tasks: NER (MasakhaNER 1.0), SC (GLUE), and QA (TyDiQA-GoldP), and scripts -- as one of the main challenges in NLP is building reliable models that can scale up to real-world applications where many scripts are often encountered. Additionally, we carry out a calibration analysis to examine the relationship between model performance and uncertainty across tasks. The performance is measured using Root Mean Square Error ($\text{RMSE} = \sqrt{\text{MSE}}$, Equation \ref{eq:mse-loss}), while uncertainty is quantified using MC standard deviation. The goal is to evaluate how well the predicted uncertainty values align with actual performance errors across the different scripts and languages. 

\begin{figure}[t]
    \centering
    \includegraphics[width=1\linewidth]{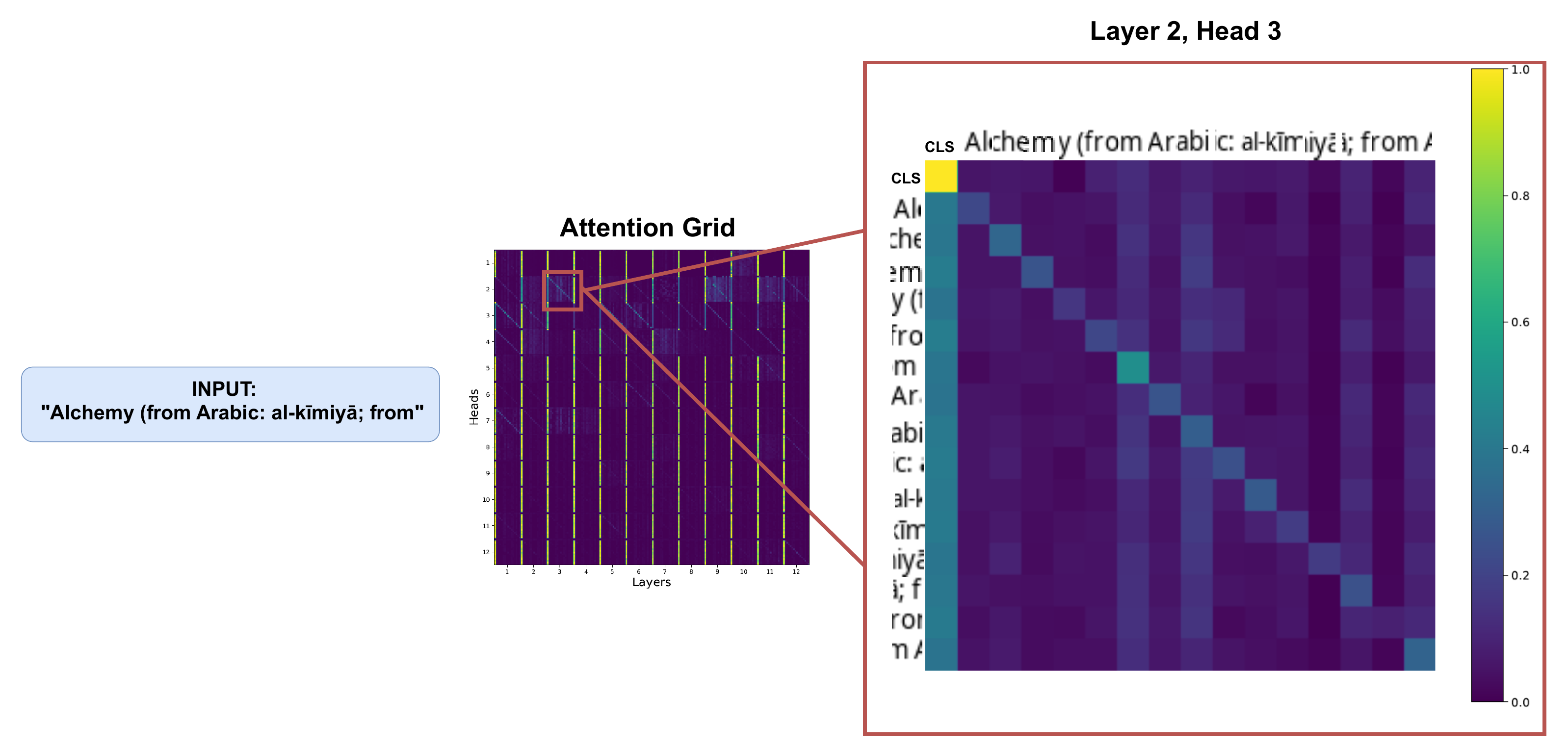}
    \caption{Model-level (attention grid) and neuron-level (layer $2$, head $3$) views of attention in the PIXEL model for a short input text from the English Wikipedia. The attention grid contains 12 attention layers with $12$ attention heads each. }
    \label{fig:attention_alchemy}
\end{figure}

\textbf{Attention Visualization}
To visualize attention in the PIXEL encoder, a square attention grid $A \in \mathbb{R}^{L \times H \times {N_\text{patches}}^2}$ is created for the encoded patches, where L is the number of attention layers and H is the number of heads in each layer. An example is presented in Figure \ref{fig:attention_alchemy}. This shows model-level attention across all layers and heads for a particular input image. Each cell $A(l, h)$ in this grid visualizes the neuron-level attention weights for a specific head $h$ and layer $l$. Then, each patch in the attention cell attends to the other patches in the sequence according to the dot product between the query (of the \textit{attender} patch) and the key (of the \textit{attended} patch). The weights are averaged over 100 Monte Carlo forward passes. Considering the increased dimensionality of the attention cell, only the first 16 patches are visualized, resulting in an image with $16 \times 16$ patches.

\textbf{Ensemble Learning} To solve the \textit{Extractive Question-Answering} task, four learner models are finetuned on each of the $9$ languages of the TyDiQA-GoldP (Section \ref{sec-data-TyDiQA}) dataset, resulting in $36$ total models. Each model is trained on the \texttt{train} split of a language in the dataset and evaluated on the \texttt{validation} split of the same language. There are four main steps to be followed to compute the final prediction for an input question. In a regular non-ensemble setting, there is only one finetuned model that dictates the output answer for each example. In the ensemble learning framework, each model $M_i$ is applied to the input question $q$ to obtain the candidate answers with corresponding confidence probability values. To reduce the pool of candidates, only the predictions that appear in all models are kept. The average confidence $\text{conf}_c$ is computed for each candidate across all models. Finally, the candidate with the highest confidence is selected. 

In the \textit{Named Entity Recognition task}, five learner models are finetuned on each of the $10$ languages of the MasakhaNER 1.0 dataset \cite{adelani2021masakhaner}, resulting in $50$ total models. Each model is trained on the \texttt{train} split of a language in the dataset and evaluated on the \texttt{test} split of the same language. The task involves assigning a label to each token from a list of $9$ predefined classes. Their predicted logits are averaged and combined into one value for each class. The final label is computed as shown in Equation \ref{eq:NER-ensemble}, where $L$ is the set of labels (classes) and $k$ is the number of models. 

\begin{equation}
    \text{label} = \arg\max_{l \in L} \left( \frac{1}{k} \sum_{i=1}^{k} \text{logits}_{i, l} \right)
    \label{eq:NER-ensemble}
\end{equation}

During the ensemble experiment, only the values of the batch size (BSZ), learning rate (LR), dropout probability (DP), and the seed are changed. For more details about the finetuning configuration and routine, refer to Tables \ref{tab:NER-parameters} and \ref{tab:QA-parameters}.

\section{Results}

\subsection{Monte Carlo Uncertainty}

\begin{figure*}[t]
  \includegraphics[width=0.5\linewidth]{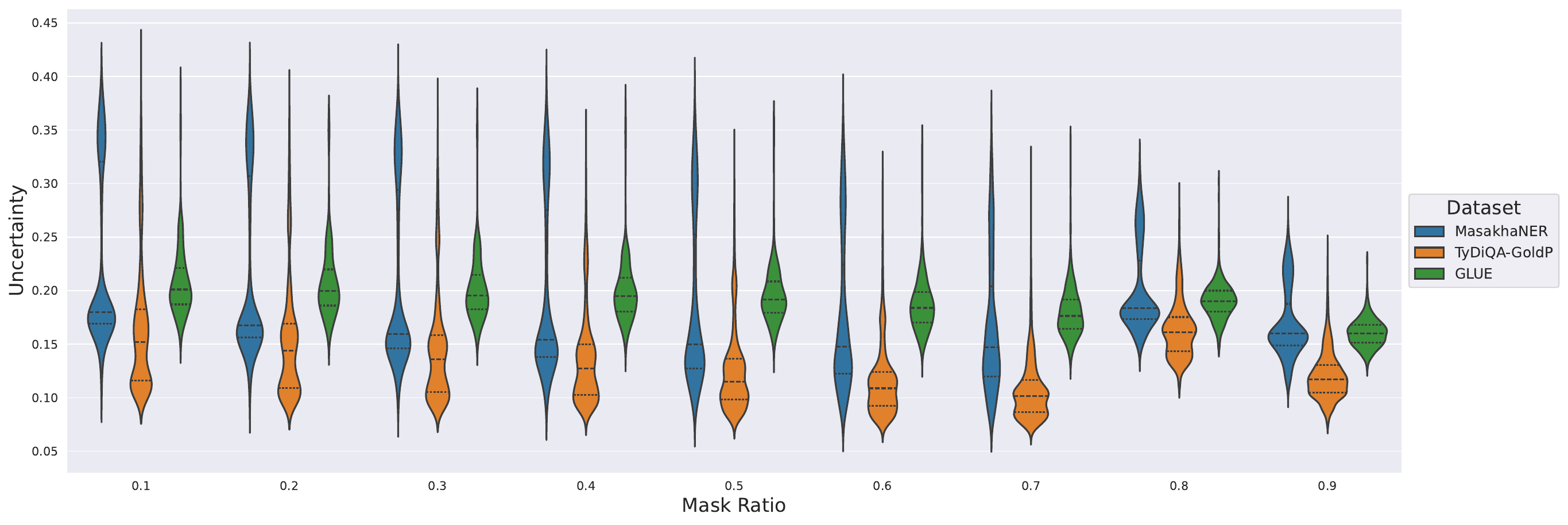} \hfill
  \includegraphics[width=0.5\linewidth]{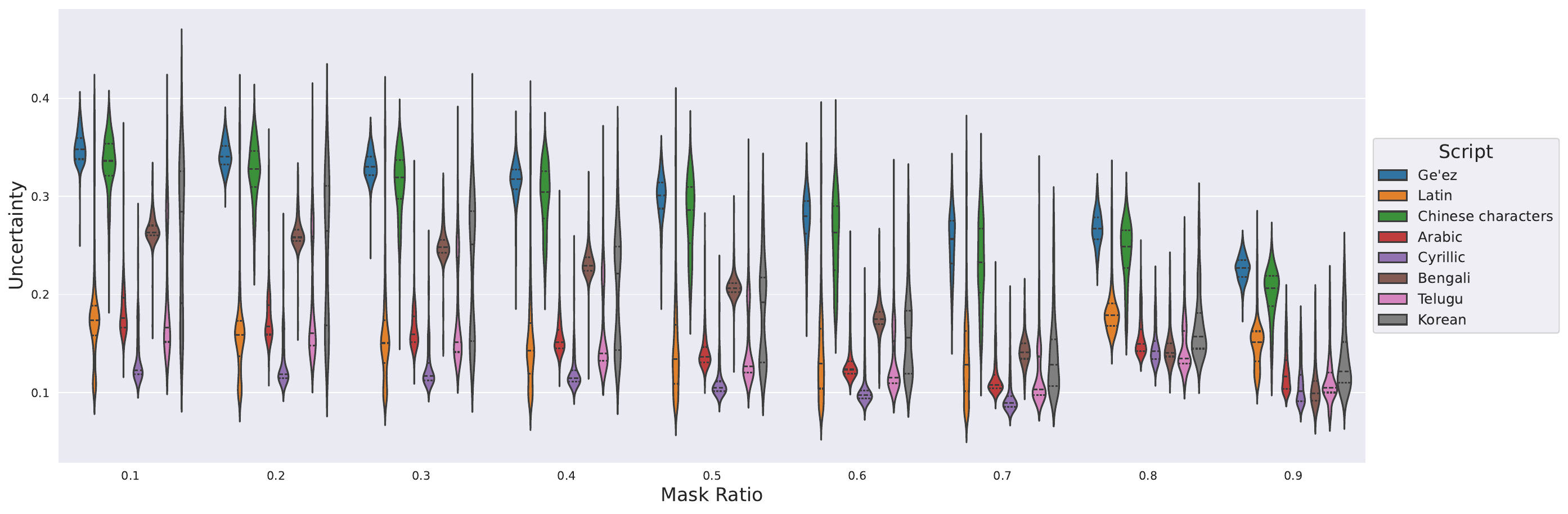}
  \caption {The distribution of the MC Uncertainty across the different datasets (left) and scripts (right) for each mask ratio value $R$.}
  \label{fig:violin-plots}
\end{figure*}

\begin{figure*}[t]
  \includegraphics[width=0.5\linewidth]{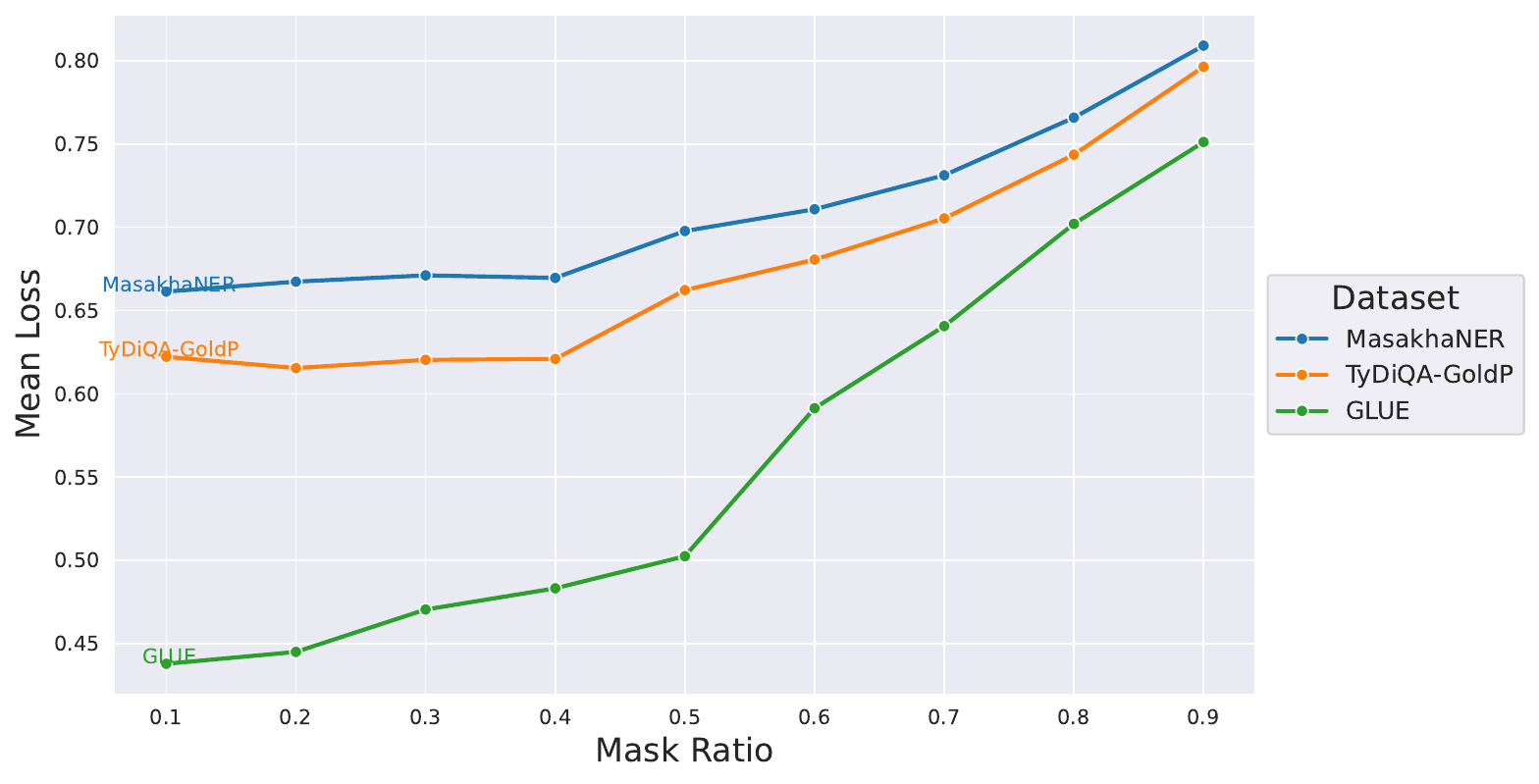} \hfill
  \includegraphics[width=0.5\linewidth]{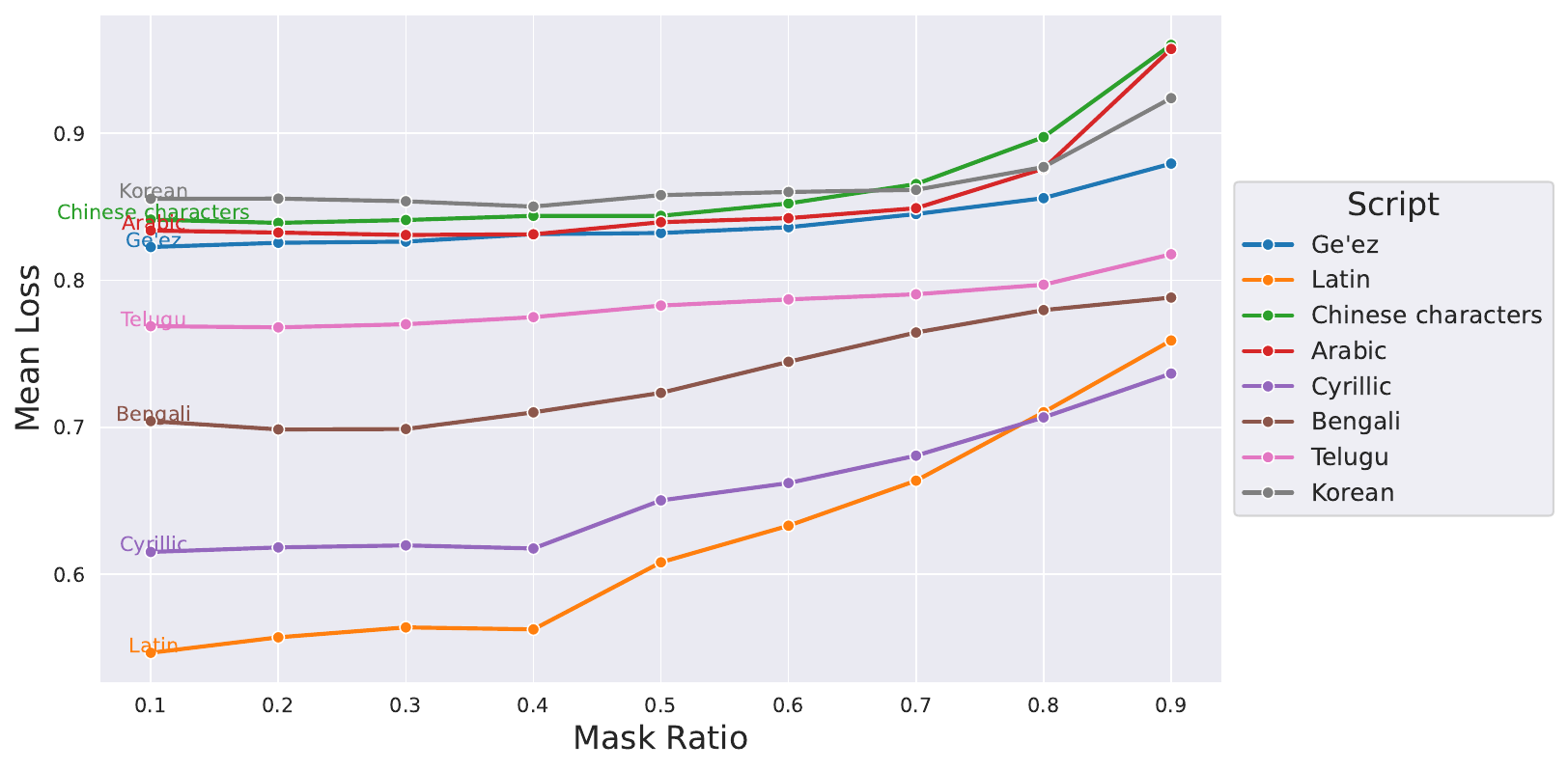}
  \caption {The MSE loss across the different datasets (left) and scripts (right) for each mask ratio value $R$.}
  \label{fig:line-plots}
\end{figure*}

\textbf{Uncertainty Across Datasets} The distribution of MC uncertainty is presented in Figure \ref{fig:violin-plots} (left), suggesting that GLUE achieves the highest overall uncertainty, which indicates that pixel-level uncertainty increases with text that has more semantic complexity, as it is the case in sentiment classification, semantic similarity or textual entailment tasks. 

In terms of the mask ratio R, the plot indicates that lower values (0.1 to 0.3) generally correspond to lower uncertainty across all datasets, hinting that less masking leads to more certain predictions. In this case, the largest part of the data is concentrated between uncertainty values of 0.15 and 0.25. As the mask ratio increases, the distribution becomes more spread out.

The results from Figure \ref{fig:line-plots} (left) indicate that the loss increases with the mask ratio. This is expected as the model was trained to reconstruct the image patches with a mask ratio of $R=0.25$. There is also a wide performance gap between the sequence classification task (GLUE) and the rest of the tasks, which can be attributed to language. The GLUE dataset contains English text, the language the PIXEL model was pretrained on, while TyDiQA-GoldP and MasakhaNER are multilingual datasets.

\textbf{Uncertainty Across Scripts} The overall trends (right) show that Ge’ez, Chinese Characters, Arabic, and Korean scripts exhibit high uncertainty (Figure \ref{fig:violin-plots}, right) and high mean loss (Figure \ref{fig:line-plots}, right), and the increase is more pronounced at mask ratios above 0.6. The Latin and Cyrillic scripts are increasing more gradually with a sharper uptick around 0.8 -- 0.9. The main script found in the pre-training datasets (English Wikipedia and the BookCorpus) is Latin, and there is a high overlap between Latin and Cyrillic characters, given that both scripts share Greek as a common ancestor. However, the uncertainty in the Cyrillic script is lower, compared to Latin. The scripts with the highest MC uncertainty are Ge’ez and Chinese Characters, both of which are visually quite distinct from the Latin script.

\begin{figure}[t]
    \centering
    \includegraphics[width=1\linewidth]{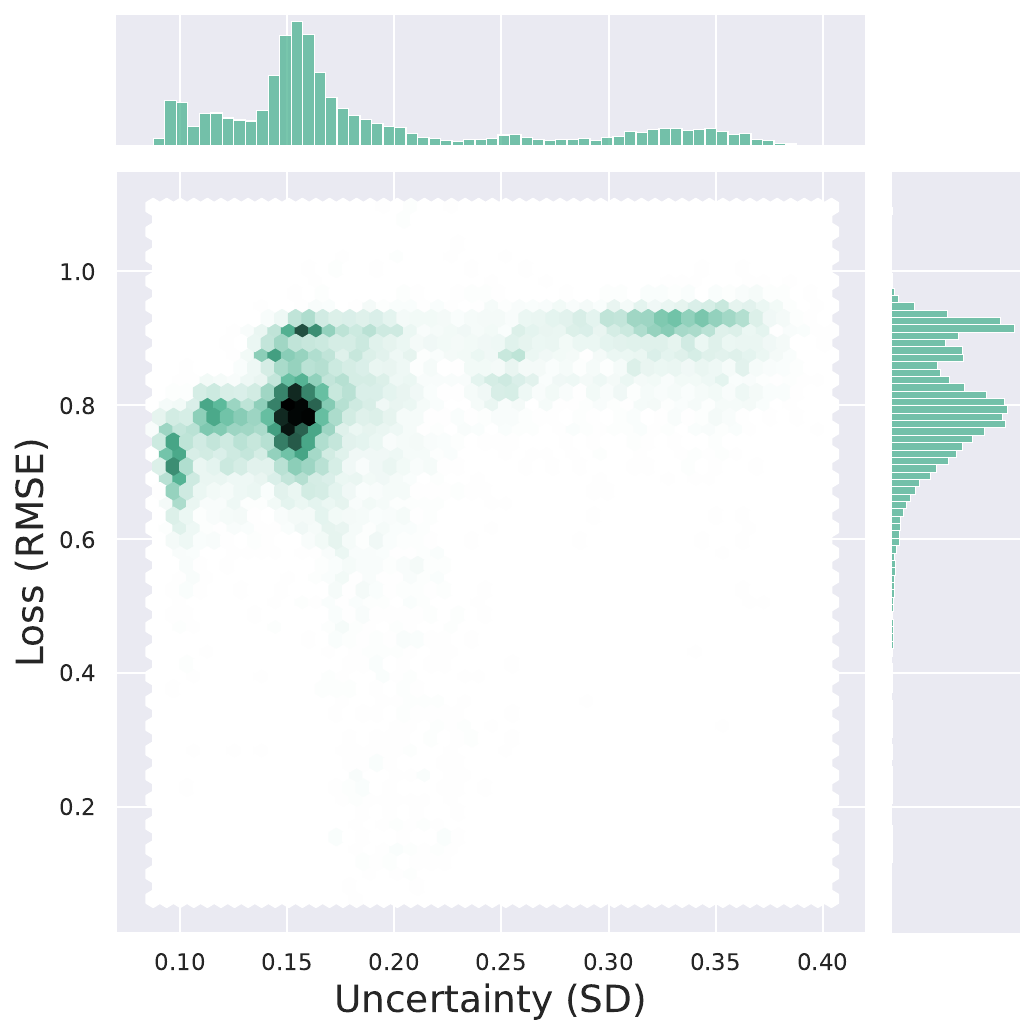}
    \caption{Calibration hexbin plot showing the RMSE loss in terms of the MC uncertainty.}
    \label{fig:calibration-overall}
\end{figure}

\textbf{Calibration Analysis} To further study the relationship between performance and uncertainty, Figure \ref{fig:calibration-overall} depicts a hexbin plot with marginal distributions, where the Root Mean Squared Error (RMSE) loss is plotted against the SD uncertainty from the MC experiments. The x-axis represents the aggregated per-image standard deviation (uncertainty) of the model after 100 Monte Carlo samples. The RMSE measures the average of the actual errors between the true pixel values and the predicted values. Inside each hexagon, the color intensity corresponds to the density of data points within that hexagon. Therefore, darker regions indicate a higher density of data points. There is a high density of points in the top left corner, which suggests that the model underestimates its performance. In other words, many examples are associated with high loss but low uncertainty. 

\begin{figure}[t]
    \centering
    \includegraphics[width=1\linewidth]{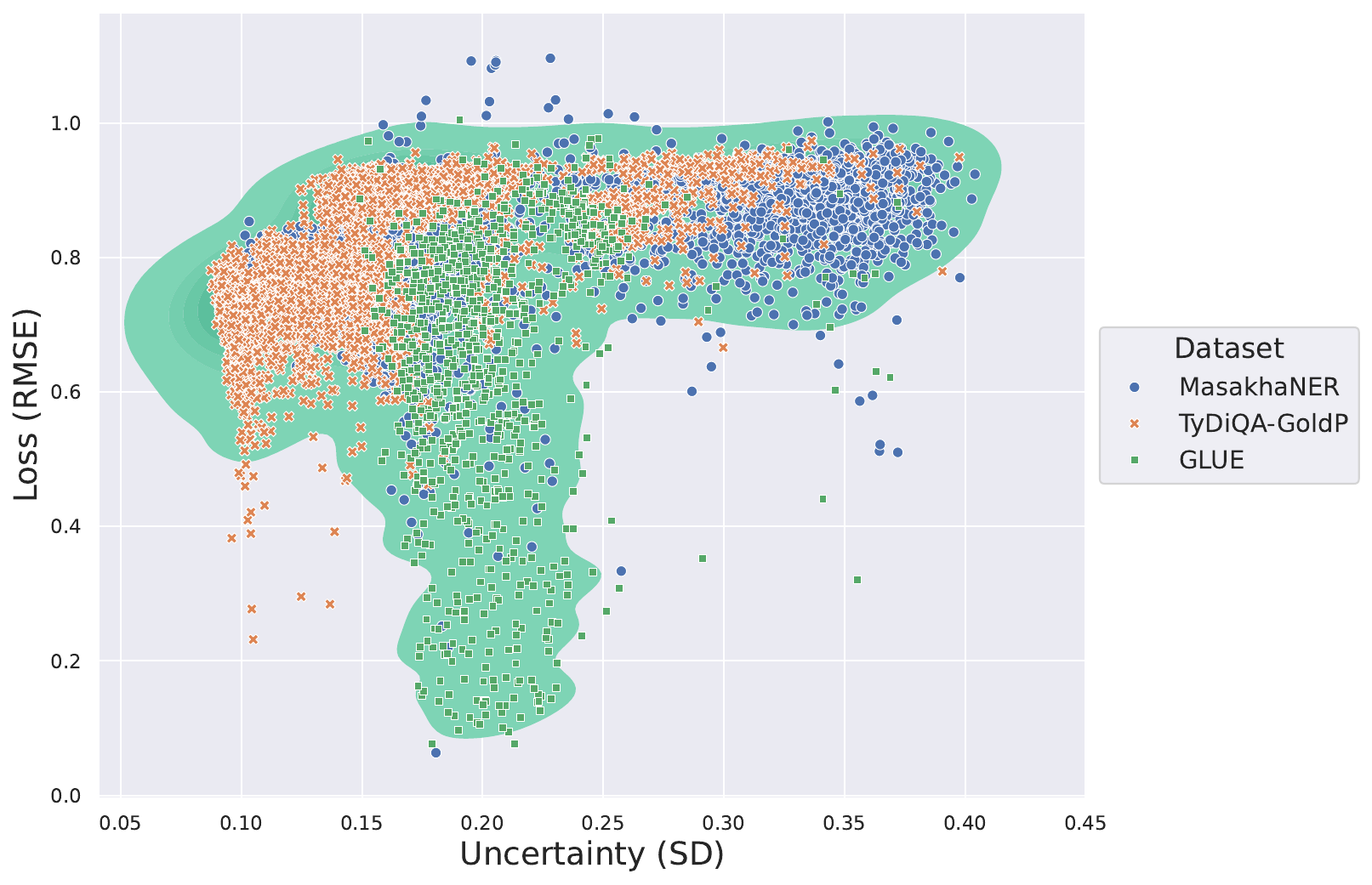}
    \caption{Calibration kernel density estimate plot showing the RMSE loss in terms of the MC uncertainty across the three datasets.}
    \label{fig:calibration-plots}
\end{figure}

The distribution of the points for all three datasets (MasakhaNER, TyDiQA-GoldP, and GLUE) is shown in the calibration plot from Figure \ref{fig:calibration-plots}. The highest level of overconfidence is associated with the question-answering task in TyDiQA-GoldP. However, there seems to be a subgroup of points for which the uncertainty is high. The points in the MaskhaNER dataset fall under the category of high uncertainty and high loss. The GLUE data is located between 0.15 and 0.3 on the uncertainty range and contains several examples showing decreased loss. While the model can be considered to be underestimating uncertainty with this group, the majority of the data still fall over the main diagonal, indicating an underestimation of uncertainty.

\textbf{Visualizing Uncertainty in Text Reconstruction}
Figure \ref{fig:MC-uncertainty-first} shows $(a)$ the original rendered English text generated with the PyGame text renderer, $(b)$ the original image overlaied with per-patch uncertainty and $(c)$ the reconstructed text overlaied with per-patch uncertainty. Bright yellow patches suggest larger variations in predictions. This can be observed in the larger masked segments of patches from the first \num{6} lines of the image, as well as in lines \num{12} and \num{15}. These segments also translate to less accurate reconstructions, as seen on the corresponding rows of the reconstructed image. On the other hand, smaller segments of patches (which appear darker in the image) are associated with lower uncertainty and are reconstructed more accurately. These patches often contain shorter sequences of letters. In terms of the mistakes, the model fails to reconstruct patches with numerals, such as \textit{20-fold}. Still, it appears to understand that the most suitable prediction given the context is a number (the model predicts \textit{20,000}). Moreover, longer and less frequent words such as \textit{implementation} and \textit{publish}, as well as punctuation marks (used in \textit{(LLMs)}) appear to produce more variation in the prediction, given the increased uncertainty.

\subsection{Attention Visualization}

\begin{figure*}[t]
  \includegraphics[width=0.48\linewidth]{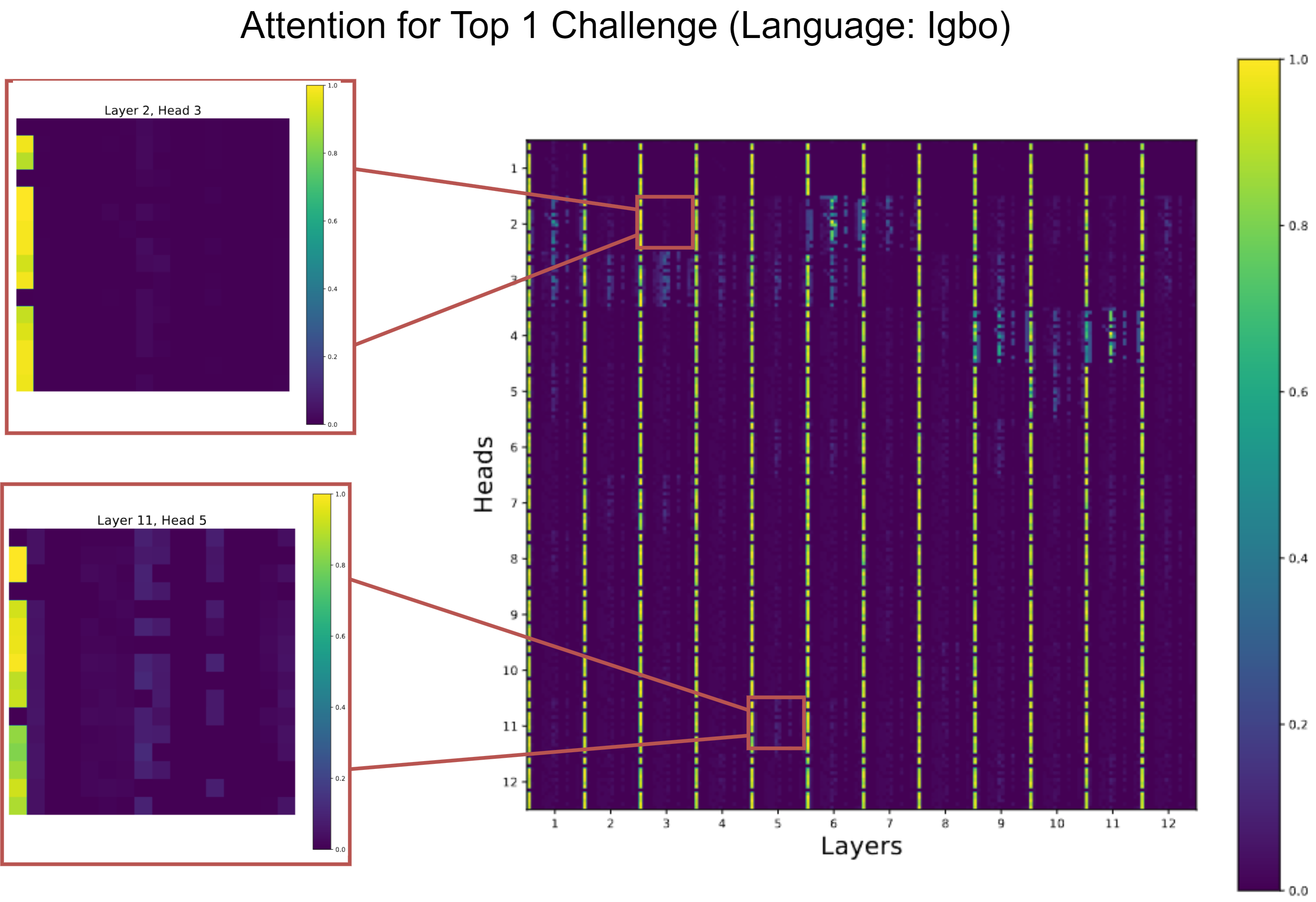} \hfill
  \includegraphics[width=0.48\linewidth]{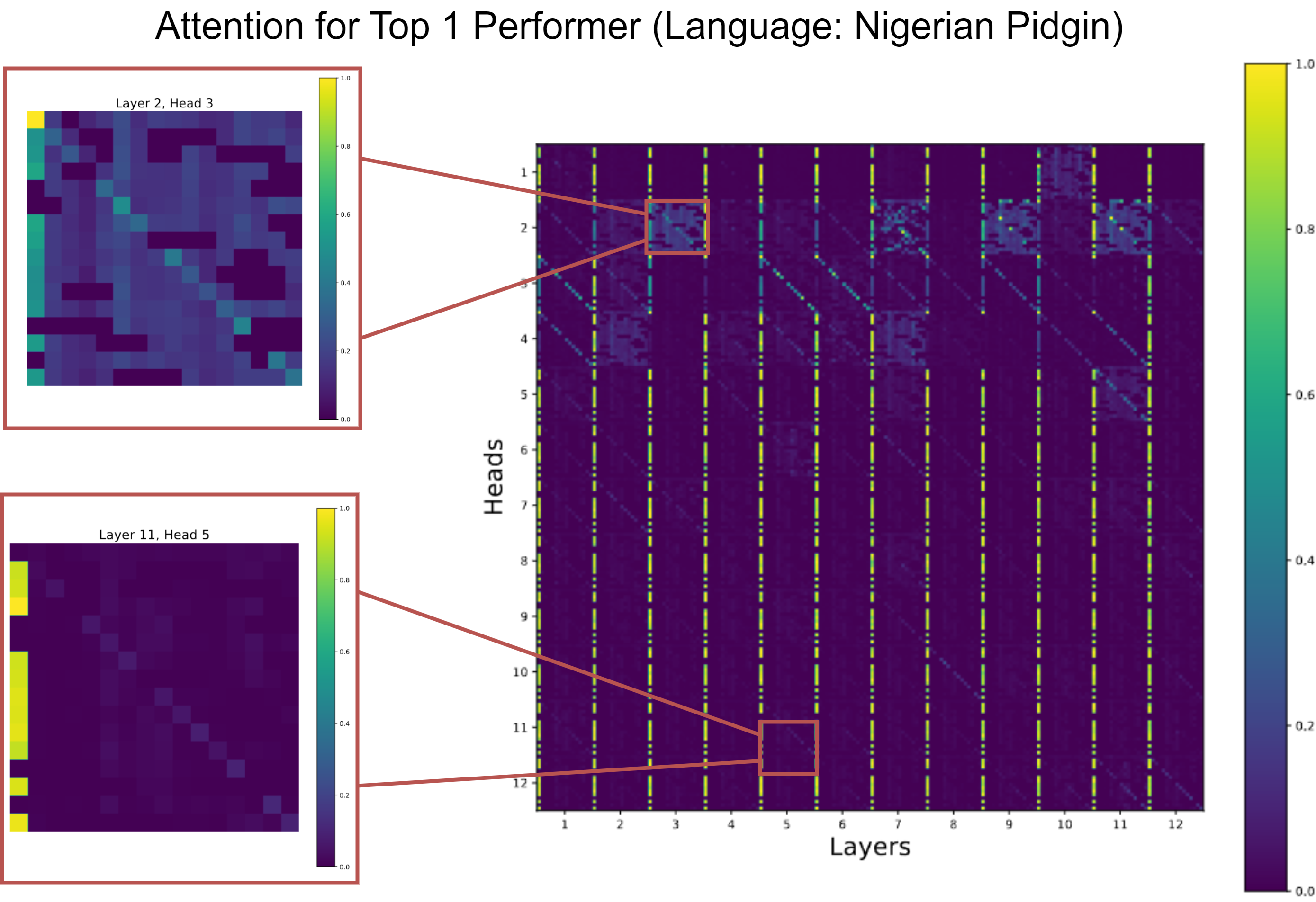}
  \caption {Model-level and neuron-level views of attention for the top 1 challenge (left, highest loss value) and performer (right, lowest loss value) in terms of the GNLL loss across all datasets.}
  \label{fig:igbo-nigerian-attention-plots}
\end{figure*}

 Each cell in the attention grids (Figure \ref{fig:igbo-nigerian-attention-plots}) shows the attention weights for the first 16 patches of a specific head $h$ and layer $l$ in the selected examples. The first four layers appear to encode the highest amount of visual information, given the high activation of the patches. Across all heads and layers of both examples, the attention weight corresponding to the CLS patch is high, as it contains the aggregate representation of the input patch sequence. There is a clear difference in the distribution of attention between the examples. The top 1 performer (Nigerian Pidgin) exhibits high activation on the diagonal at the neuron level, meaning that patches are attending to themselves, possibly to retain positional and contextual information. The Igbo example does not show the same pattern, rather a subset of dominant patches attend to the remaining ones.

\subsection{Ensemble Learning}

\begin{figure}[t]
    \centering
    \includegraphics[width=1\linewidth]{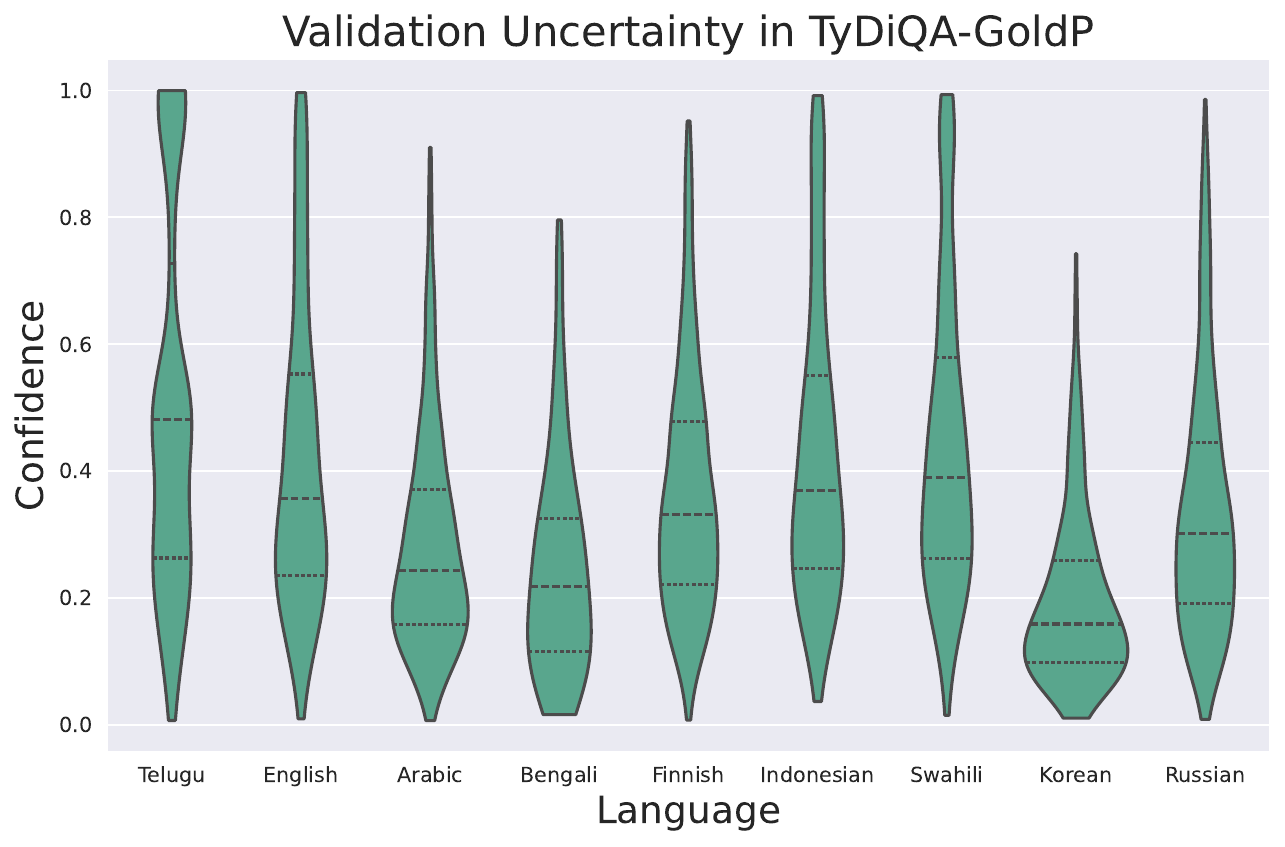}
    \caption{Confidence distribution across all languages in the TyDiQA-GoldP dataset for the ensemble model.}
    \label{fig:QA-violin}
\end{figure}

\textbf{Extractive Question Answering}
The results of the ensemble QA model are presented in Table \ref{tab:QA-f1score}, which shows the weighted F1 score across all languages in the TyDiQA-GoldP dataset. These findings are compared with the results obtain by \citet{rust2022language}, following the same experimental setting. Overall, the ensemble learning method improves the performance in the extractive QA task for $6$ out of the $8$ languages. The average F1 score (excluding the $ENG$ data) for the ensemble configuration is higher with $1.7$ points than in the case of the regular PIXEL model. In terms of the individual languages, there is a high improvement for Indonesian ($4.3$ points), Russian ($2.8$ points), and Arabic ($2.2$ points), suggesting that combining multiple learners can improve performance regardless of script.

Figure \ref{fig:QA-violin} presents the confidence distribution of the best answers in the ensemble model for all languages in the dataset. In general, the confidence is in the range $0.2 - 0.4$ across the majority of languages, with some distributions indicating slightly higher confidence, as in the case of Finnish, Indonesian, and Swahili. Lower confidence values can be seen in Korean and Bengali. These observations are in line with the previous findings on performance.

\begin{table*}[t]
\centering
\begin{tabular}{ccccccccccc}
\hline
 & ARA & BEN & FIN & IND & KOR & RUS & SWA & TEL & ENG & \textbf{AVG} \\
\hline
\textbf{PIXEL}  & 57.3 & \textbf{36.3} & 58.3 & 63.6 & 26.1 & 50.5 & 65.9 & 63.4 & 61.7 & 52.3 \\
\textbf{Ensemble} & \textbf{59.5} & 35.1 & \textbf{59.6} & \textbf{67.3} & \textbf{27.1} & \textbf{53.3} & \textbf{67.1} & 63.4 & \textbf{62.1} & \textbf{54.0} \\
\hline
\end{tabular}
\caption{The results of the QA task. The ensemble learning model finetuned on the TyDiQA-GoldP dataset is compared with the values reported by \cite{rust2022language}. The metric shown is the F1 score, computed on the \texttt{validation} split of the data. The $AVG$ score excludes $ENG$, as required \citep{clark2020tydi}.}
\label{tab:QA-f1score}
\end{table*}

\textbf{Named Entity Recognition} The results of the ensemble NER model are presented in Table \ref{tab:NER-f1score}, showing the weighted F1 score across the MasakhaNER 1.0 dataset. Due to hardware limitations at runtime, the \textit{ENG} data is not included. For comparison, the results are shown against the values obtained by \citet{rust2022language}. In general, ensemble learning improves the performance significantly for all $9$ languages, resulting in scores higher than $90$. This is also the case for languages that were previously associated with a low score, such as Amharic (\textit{AMH}). The F1 score gap is $24.3$ points in favour of the ensemble method, suggesting that ensemble learning improves the comprehension of long-term dependencies in NER tasks.

\begin{table*}[t]
\centering
\begin{tabular}{cccccccccccc}
\hline
& AMH & HAU & IBO & KIN & LUG & LUO & PCM & SWA & WOL & YOR & \textbf{AVG} \\
\hline
\textbf{PIXEL} & 47.7 & 82.4 & 79.9 & 64.2 & 76.5 & 66.6 & 78.7 & 79.8 & 59.7 & 70.7 & 70.7  \\
\textbf{Ensemble} &\textbf{ 90.2} & \textbf{97.1} & \textbf{96.1} & \textbf{93.9} & \textbf{95.5} & \textbf{93.1} & \textbf{97.1} & \textbf{96.1} & \textbf{95.8} & \textbf{95.2} & \textbf{95} \\
\hline
\end{tabular}
\caption{The results of the NER task. The ensemble learning model finetuned on the MasakhaNER 1.0 dataset is compared with the values reported by \cite{rust2022language}. The metric shown is the F1 score, computed on the \texttt{test} split of the data.}
\label{tab:NER-f1score}
\end{table*}

\section{Discussion}

This work showed that it is possible to integrate uncertainty quantification methods and measure calibration in the context of visual text models. These methods include Monte Carlo Dropout at the patch level, with the observation that more work should be directed towards finding more effective ways of aggregating and visualizing uncertainty across longer patch sequences. Attention based methods can also be used to gain insights into how these models encode information, but there remains the debate about whether or not attention counts as an explanation \citep{bibal2022attention}. Still, this debate falls outside the scope of this research. Ensemble learning with a low number of individual learners can also be used successfully to improve both performance and confidence. 

The results in the MC Uncertainty experiment generally indicate high uncertainty for a high mask ratio. Still, the most optimal value is a mask ratio of $50\%$, representing a reasonable trade-off between uncertainty and loss. 

Scripts such as Latin are less uncertain, indicating that multilingual pretraining is necessary. Instead of language, one can focus on introducing a new script, as evidence suggests that there exists knowledge transfer between scripts like Latin and Cyrillic. For example, finetuning on one language such as Chinese might benefit performance in other languages like Korean or Amharic. This approach is more robust than traditional LLMs, where the transfer of learning happens under stricter conditions, for instance when languages share syntactic structures or when there is a significant overlap between vocabularies.

Ensemble learning can be applied successfully to improve performance and calibration in pixel-based language models. The evaluation shows higher F1 scores for 17 of the 19 tested languages across two tasks. The models become more robust and can overcome individual weaknesses by aggregating predictions from multiple learners using hyperparameter tuning. Additionally, ensemble learning improves calibration through better error diversification and data representation.

\section{Conclusions and Future Work}

The findings of this study indicate that pixel-based language models represent a viable and lightweight solution to traditional language modeling, even for tasks that require semantic understanding of text. Their reliability and explainability can also be improved through uncertainty quantification methods, as shown during the experiments. Future research should focus on perfecting the existing techniques and exploring new ways of understanding the inner workings of models that enccde text as visual representation.

One point to be explored in future works on text reconstruction is the idea of pixels-as-tokens in the context of the Pixel Transformer (PiT) model, introduced by \cite{nguyen2024image}. Instead of training the model to perform patch reconstruction, PiT treats each pixel as a token and the reconstruction happens at the pixel level. Evidence suggests that this method completely removes locality as in inductive bias. This can potentially improve long-term context comprehension in the proposed approach, as the current findings indicate that the reconstruction of characters depends on neighboring pixels. Additionally, the finetuning pipeline can be expanded to more complex semantic tasks, such as summarization, open-ended question answering where the answer is not always explicitly mentioned in the context, and text generation (\citet{li2023renderdiffusion} introduced a new method for text generation using GlyphDiffusion). To improve model calibration, post-hoc methods like temperature scaling can be used either separately or in combination with Monte Carlo \citep{laves2019well}. During pretraining, the Cross-Entropy loss can be replaced by the Focal Loss, which is effective in calibration models trained on imbalanced datasets \citep{wang2022calibrating}. 

\section*{Ethical Considerations}

The aim of this study is to shed light on how pixel-based models encode uncertainty. We consider that an explainability analysis should be a prerequisite for any new language model, as this increases users' trust that the technology works as intended and it is not harmful.

In order for this research to exist, we made use of the pretrained PIXEL model provided by \citet{rust2022language}. One of the datasets that PIXEL has been pretrained on is the BookCorpus \cite{zhu2015aligning} which is well-known for its problematic content and copyright violantions \cite{bandy2021addressingdocumentationdebtmachine}. BookCorpus contains books self-published by authors, which did not explicitly consent to including their books in a LLM training dataset, and were not compensated in any way. Moreover, many books contain copyright restrictions which forbid the redistribution of content. Senstive content has also been identified in the data, such as books marked for adult audiences, containing terms and phrases associated with gender discrimination. We acknowledge that by using models trained on problematic data, we risk to further propagate biases. However, these models and datasets are very popular and they cannot be ignored. For this reason, we consider that studying how they work and attempting to explain and interpret them is a goal worth pursuing.  

Our paper has a strong focus on language variety, as we explore uncertainty across \num{18} languages. However, the majority of our fine-tuning data comes from English (as seen in Figure \ref{fig:MC_languages_hist} from Appendix \ref{sec:appendix-data-details}). This leads to lower performance and less accurate representation in low-resource languages. Once again, this issue boils down to the data available for LLM training, which should ideally be more balanced and representative across diverse linguistic contexts.

\section*{Code}
\label{sec:appendix-code-details}
We provide the complete implementation for running our experiments on Github, at \url{https://github.com/stefania-radu/pixel-semantic}.

\bibliography{custom}

\appendix
\onecolumn
\newpage

\section{Limitations}

Some limitations of this method include the hardware and training time required to train multiple models. Nevertheless, PIXEL has $20\%$ fewer parameters than BERT, so an ensemble of PIXEL models remains less complex than the BERT variant and significantly more lightweight than models like GPT.

The current study is subject to several limitations. Firstly, the way uncertainty is computed at the image level during the MC experiments can be more reliable. At the moment, uncertainty is averaged across all pixels in an image. However, this does not account for the difference in span length, as some sequences of patches are longer than others. Quantifying uncertainty as an average for each span length in the image could bring more insights into how the model encodes long-term dependencies. Secondly, the information in the attention plots should be aggregated so that all patches are visible at once, while keeping a reasonable image size. Using the current method, visualizing all 256 patches across the 144 attention structures would result in a very large and difficult to interpret image. Regarding the calibration analysis, it is not completely clear that the two measurements of performance (loss vs. MC uncertainty during the pretraining stage and F1 score vs. confidence during finetuning) are quantifying the same underlying metric. For this reason, additional testing should be performed to establish the exact effect size of ensemble learning on model calibration. Moreover, more insights are necessary to establish the trade-off between computational cost, environmental impact and performance gains when training an ensemble of learners compared to a single model.

While it is possible to visualize the attention mechanism in pixel-based language models, there are some comments to be made about this. Unlike traditional language models like BERT where each token represents a meaningful unit and the relationship between two tokens can be understood intuitively, the patches in pixel-based language models cannot be mapped back to text chunks. This makes it more challenging to interpret how attention is paid to the different patches and what are the implications of these connections in the context of the entire model. Moreover, given the large number of attention structures and the image dimensions, visualizing attention for all patches simultaneously becomes very difficult.

\section{Data Details}
\label{sec:appendix-data-details}

\begin{figure}[h!]
    \centering
    \includegraphics[width=0.8\linewidth]{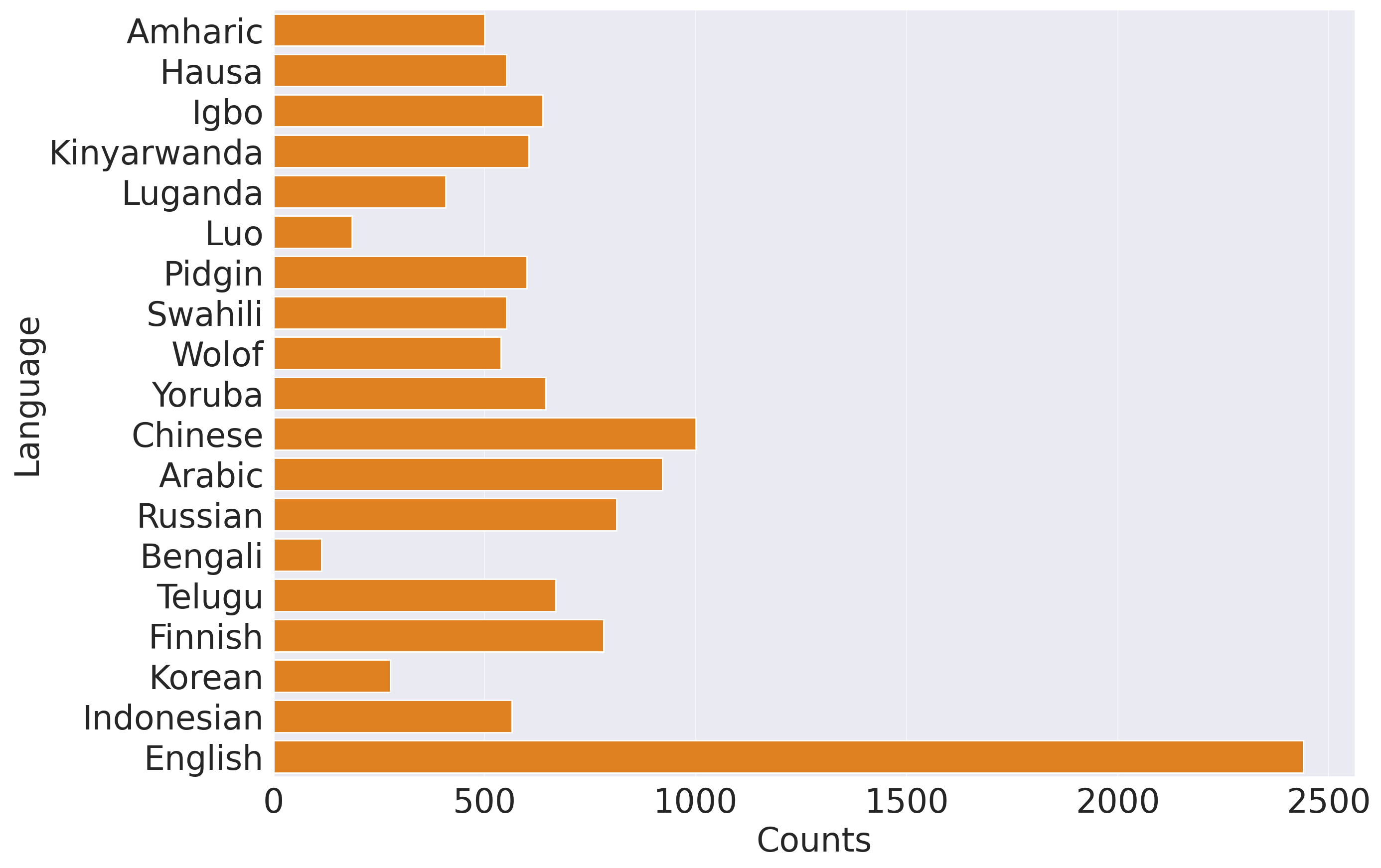}
    \caption{Distribution of languages used throughout the experiments.}
\label{fig:MC_languages_hist}
\end{figure}

\begin{table*}[h!]
\centering
\begin{tabular}{llllP{2.5cm}P{2.5cm}}
\hline
\textbf{Language} & \textbf{ISO 639-3} & \textbf{Language Family} & \textbf{Script} \\ \hline
Amharic    & AMH & Afro-Asiatic   & Ge'ez         \\ 
Arabic     & ARA & Afro-Asiatic   & Arabic       \\ 
Bengali    & BEN & Indo-European  & Bengali       \\ 
English    & ENG & Indo-European  & Latin         \\ 
Finnish    & FIN & Uralic         & Latin         \\ 
Hausa      & HAU & Afro-Asiatic   & Latin         \\ 
Igbo       & IBO & Niger-Congo    & Latin       \\ 
Indonesian & IND & Austronesian   & Latin       \\ 
Kinyarwanda& KIN & Niger-Congo    & Latin        \\ 
Korean     & KOR & Koreanic       & Korean       \\
Luganda    & LUG & Niger-Congo    & Latin         \\ 
Naija Pidgin & PCM & English Creole & Latin     \\ 
Russian    & RUS & Indo-European  & Cyrillic      \\ 
Swahili    & SWA & Niger-Congo    & Latin         \\ 
Telugu     & TEL & Dravidian      & Telugu        \\ 
Wolof      & WOL & Niger-Congo    & Latin         \\ 
Yorùbá     & YOR & Niger-Congo    & Latin        \\ \hline
\end{tabular}
\caption{An overview of languages used during the experiments. The original PIXEL model is pretrained on English only.}
\label{tab:overview-languages}
\end{table*}

\section{Experiments Details}
\label{sec:appendix-experiments}

\begin{table*}[h!]
\centering
\begin{tabular}{p{2cm}p{4.8cm}p{5cm}p{3cm}}
\hline
\textbf{Experiment} & \textbf{Data} & \textbf{Hyperparameters} & \textbf{Metrics} \\ \hline

MCU Tasks & NER (MasakhaNER 1.0), SC (GLUE), QA (TyDiQA-GoldP) & $R \in \{0.1, 0.2, \ldots, 0.9\}$, $S \in \{1, 2, \ldots, 6\}$, $W = \{0, 0, \ldots, 0, 1\}, |W| = |S|$ & MSE \newline GNLL \newline Uncertainty ($\bar{\sigma}$) \\ \hline

MCU Scripts & Latin, Ge'ez, Chinese Characters, Arabic, Cyrillic, Bengali, Telugu,  Korean & $R \in \{0.1, 0.2, \ldots, 0.9\}$, $S \in \{1, 2, \ldots, 6\}$, $W = \{0, 0, \ldots, 0, 1\}, |W| = |S|$ & MSE \newline GNLL \newline Uncertainty ($\bar{\sigma}$) \\ \hline

VU & Nigerian Pidgin, Igbo & $R = 0.25$, $S = 6$, $W = \{0.2, 0.4, 0.6, 0.8, 0.9, 1\}$ & GNLL \newline Uncertainty ($\bar{\sigma}$) \\ \hline

CA & NER (MasakhaNER 1.0), SC (GLUE), QA (TyDiQA-GoldP) & $R = 0.25$, $S = 6$, $W = \{0.2, 0.4, 0.6, 0.8, 0.9, 1\}$ & RMSE \newline Uncertainty ($\bar{\sigma}$) \\ \hline

\end{tabular}
\caption{Overview of the MC Uncertainty experiments. MCU = Monte Carlo Uncertainty; VU = Visualizing Uncertainty; CA = Calibration Analysis.}
\label{tab:MC-experiments}
\end{table*}

\begin{figure*}[h!]
  \includegraphics[width=8cm]{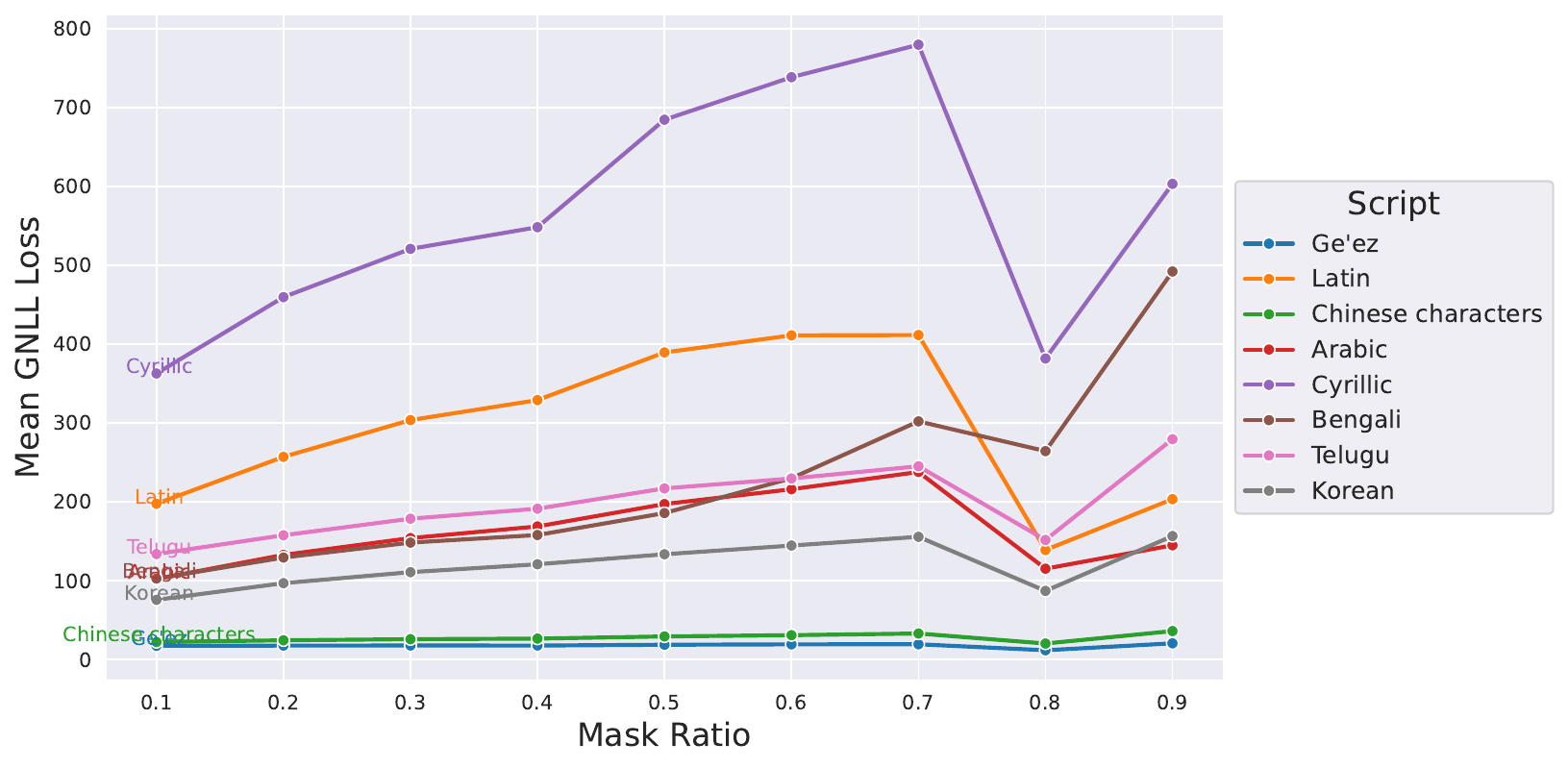} \hfill
  \includegraphics[width=8cm]{plots/Mask_Loss_scripts_line_plot.pdf}
  \caption {Mean MSE Loss (left) and GNLL Loss (right) across the different scripts for each mask ratio value $R$.}
  \label{fig:attention-plots2}
\end{figure*}

\begin{algorithm*}[h]
\caption{Patch-level Uncertainty with MC Dropout}
\label{alg:MC-uncertainty}
\begin{algorithmic}[1]
\Require Rendered image $I$, model $M$, \# MC samples $N_\text{MC}=100$, dropout rate $p=0.1$, patch size $P=16$
\Ensure Uncertainty map $U$

\Statex

\State Activate dropout in $M$

\For{$i \in \{1, \ldots, N\}$}
    \State $P_i \leftarrow M(I, p)$ \Comment{Compute predictions $P$ with dropout}
\EndFor

\State Initialize $\mu$ and $\sigma$ with the shape of $I$

\For{each pixel $(x, y)$}
    \State $\mu(x, y) \leftarrow \frac{1}{N} \sum_{i=1}^{N} P_i(x, y)$
    \State $\sigma(x, y) \leftarrow \sqrt{\frac{1}{N} \sum_{i=1}^{N} (P_i(x, y) - \mu(x, y))^2}$
\EndFor

\State Initialize $U$ with the shape of $I$

\For{each patch $(i, j)$ in $\sigma$}
    \State $\sigma_{\text{patch}} \leftarrow \frac{1}{P^2} \sum_{x=i}^{i+P-1} \sum_{y=j}^{j+P-1} \sigma(x, y)$  \Comment{Compute $\sigma$ per patch}
    
    \For {$(x, y) \in \{(i, j), \ldots, (i+P-1, j+P-1)\}$}
        \State $U(x, y) \gets \sigma_{\text{patch}}$  \Comment{ Assign  $\sigma_{\text{patch}}$ to all pixels in the patch}
    \EndFor
\EndFor

\State \textbf{return} $U$
\end{algorithmic}
\end{algorithm*}

\begin{figure*}[h]
    \centering
    \includegraphics[width=0.8\linewidth]{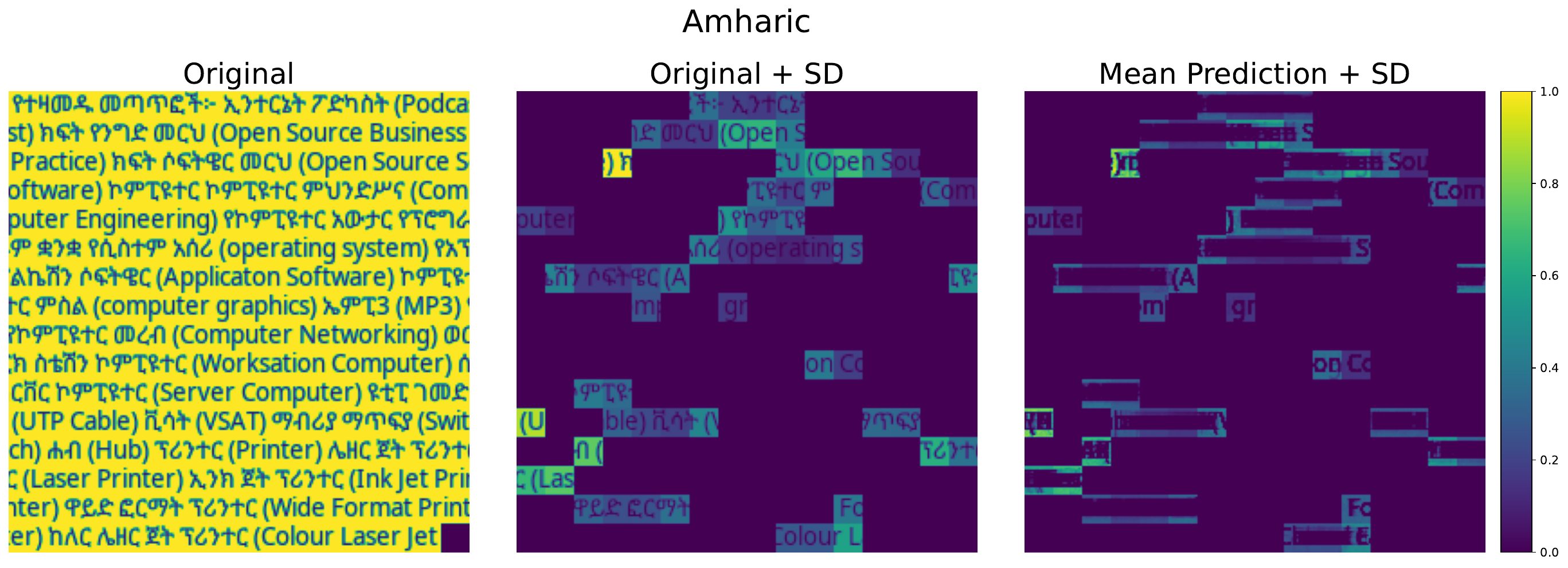}
     \includegraphics[width=0.8\linewidth]{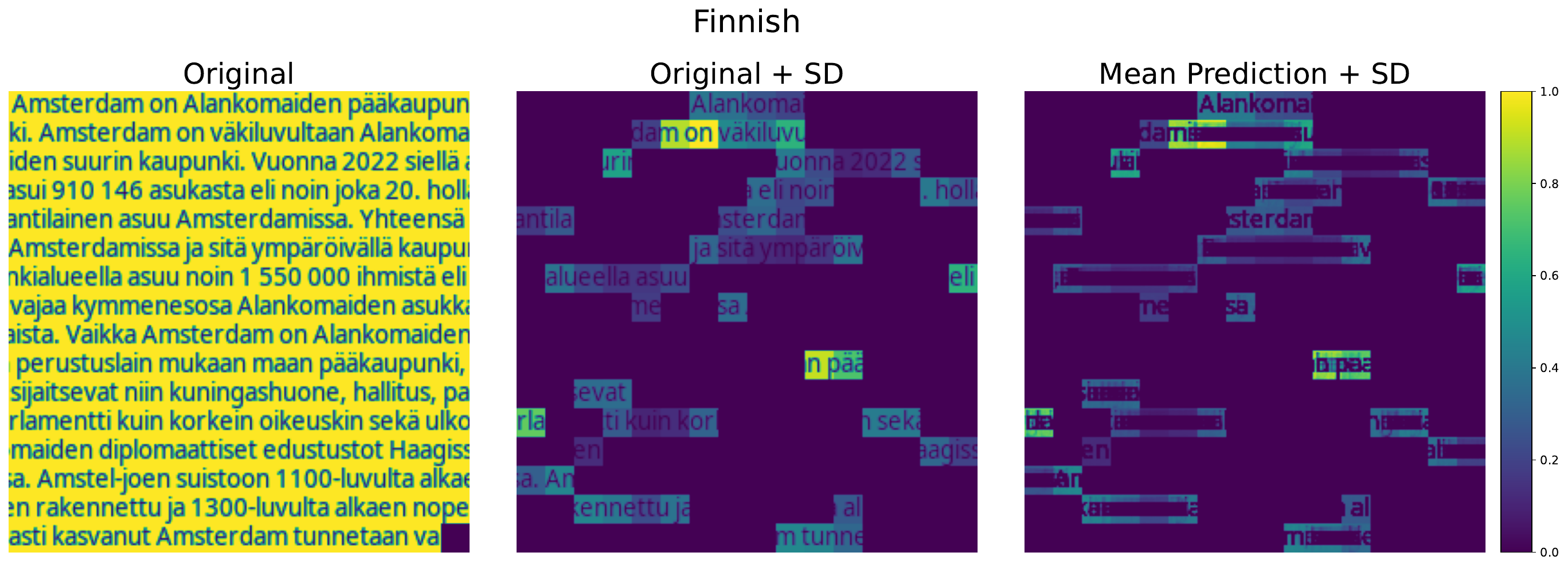}
     \includegraphics[width=0.8\linewidth]{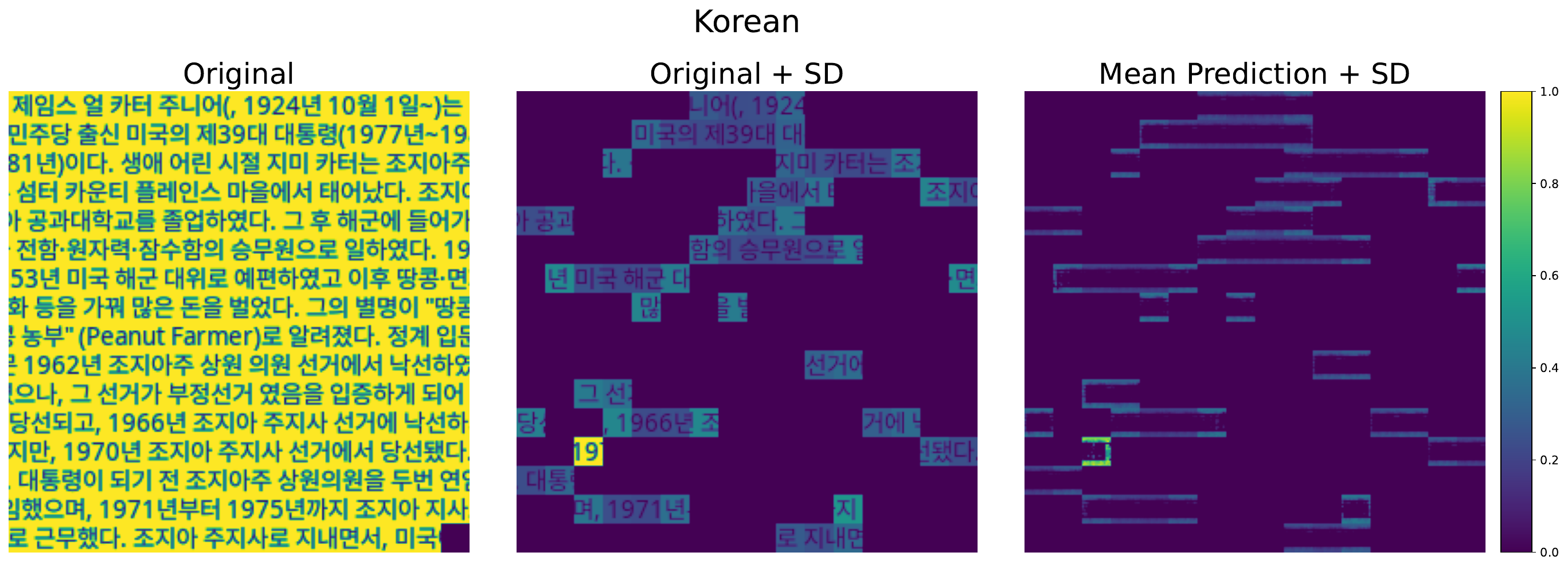}
     \includegraphics[width=0.8\linewidth]{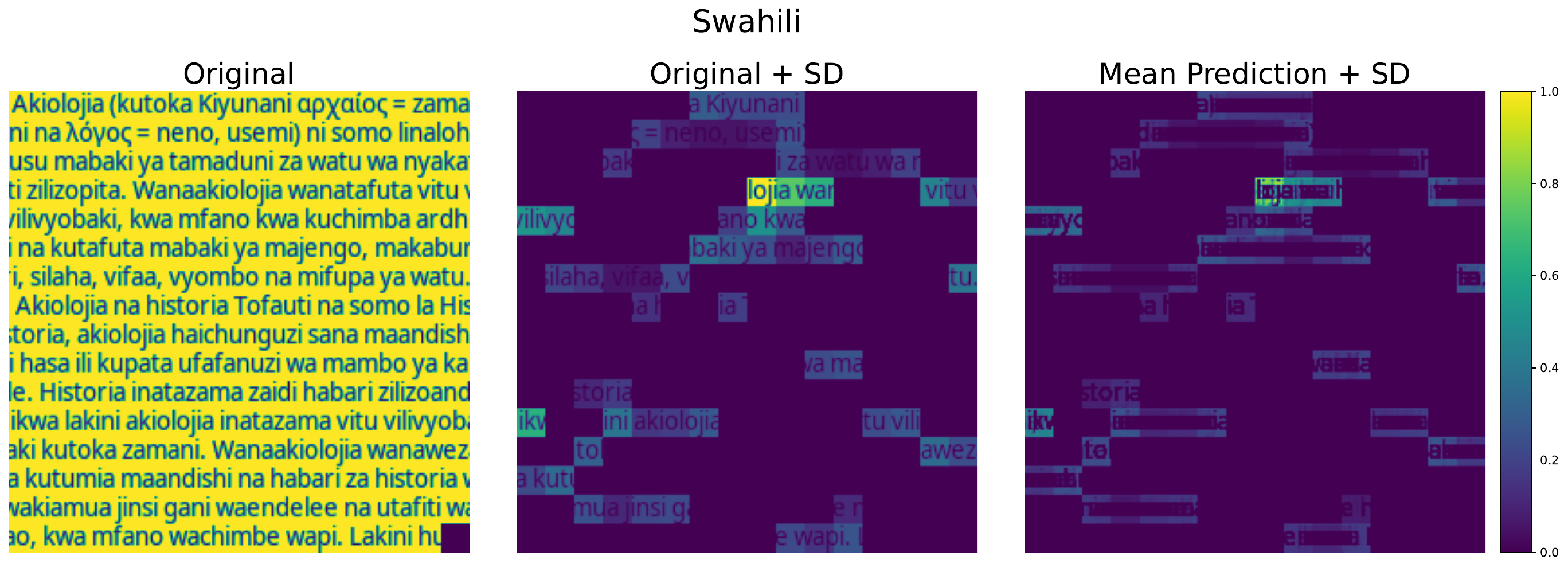}
     \includegraphics[width=0.8\linewidth]{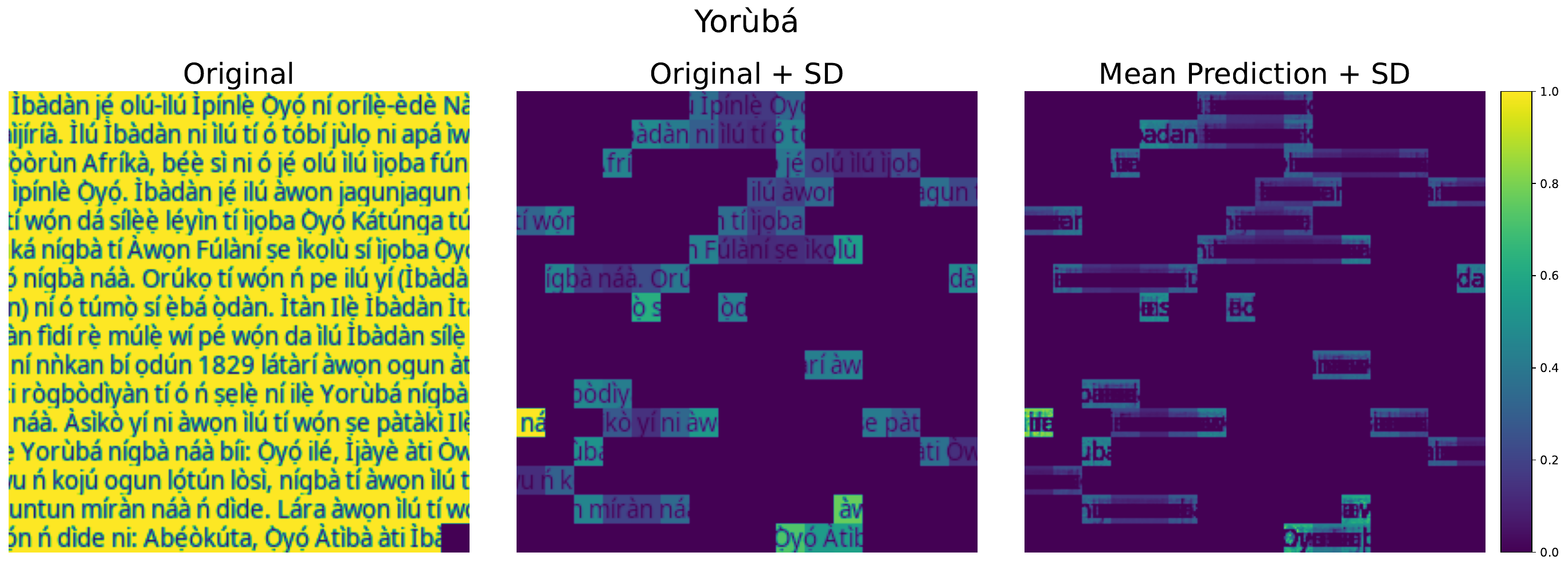}
    \caption{Examples of uncertainty quantification at the patch-level for various languages.}
    \label{fig:uncert_examples_1}
\end{figure*}

\begin{table*}[h!]
\centering
\begin{tabular}{>{\raggedright\arraybackslash}p{7cm} >{\raggedright\arraybackslash}p{8cm}}
\toprule
\textbf{Parameter} & \textbf{Value} \\ \midrule
\textbf{Common Parameters} &  \\ \midrule
Dataset name & tydiqa \\
Dataset config name & secondary\_task \\
Sequence length & 400 \\
Stride & 160 \\
Question max length & 128 \\
Gradient accumulation steps & 1 \\
Max steps & 20000 \\
Number of train epochs & 10 \\
Early stopping & True \\
Early stopping patience & 5 \\
Evaluation metric & $F1 = \frac{2 \times \text{TP}}{2 \times \text{TP} + \text{FP} + \text{FN}}$ \\
Doc stride & 160 \\
Number of best predictions & 20 \\ \midrule
\textbf{Model 1} &  \\ \midrule
Batch size & 32 \\
Learning rate & $7 \times 10^{-4}$ \\
Dropout probability & 0.15 \\
Seed & 101 \\ \midrule
\textbf{Model 2} &  \\ \midrule
Batch size & 16 \\
Learning rate & $7 \times 10^{-5}$ \\
Dropout probability & 0.15 \\
Seed & 102 \\ \midrule
\textbf{Model 3} & \\ \midrule
Batch size & 8 \\
Learning rate & $7 \times 10^{-5}$ \\
Dropout probability & 0.05 \\
Seed & 103 \\ \midrule
\textbf{Model 4} &  \\ \midrule
Batch size & 32 \\
Learning rate & $7 \times 10^{-6}$ \\
Dropout probability & 0.1 \\
Seed & 104 \\
\bottomrule
\end{tabular}
\caption{The finetuning configuration of the QA models, including the common parameters and those changed among the $4$ learners.}
\label{tab:QA-parameters}
\end{table*}

\begin{table*}[h!]
\centering
\begin{tabular}{>{\raggedright\arraybackslash}p{7cm} >{\raggedright\arraybackslash}p{8cm}}
\toprule
\textbf{Parameter} & \textbf{Value} \\ \midrule
\textbf{Common Parameters} &  \\ \midrule
Dataset name & masakhane-ner \\
Sequence length & 196 \\
Gradient accumulation steps & 1 \\
Max steps & 15000 \\
Number of train epochs & 10 \\
Early stopping & True \\
Early stopping patience & 5 \\
Evaluation metric & $F1 = \frac{2 \times \text{TP}}{2 \times \text{TP} + \text{FP} + \text{FN}}$ \\ \midrule
\textbf{Model 1} & \\ \midrule
Batch size & 64 \\
Learning rate & $5 \times 10^{-5}$ \\
Dropout probability & 0.1 \\
Seed & 100 \\ \midrule
\textbf{Model 2} &  \\ \midrule
Batch size & 64 \\
Learning rate & $5 \times 10^{-6}$ \\
Dropout probability & 0.2 \\
Seed & 101 \\ \midrule
\textbf{Model 3} &  \\ \midrule
Batch size & 32 \\
Learning rate & $5 \times 10^{-5}$ \\
Dropout probability & 0.1 \\
Seed & 102 \\ \midrule
\textbf{Model 4} & \\ \midrule
Batch size & 32 \\
Learning rate & $5 \times 10^{-6}$ \\
Dropout probability & 0.1 \\
Seed & 103 \\ \midrule
\textbf{Model 5} &  \\ \midrule
Batch size & 16 \\
Learning rate & $5 \times 10^{-5}$ \\
Dropout probability & 0.2 \\
Seed & 104 \\
\bottomrule
\end{tabular}
\caption{The finetuning configuration of the NER models, including the common parameters and those changed among the $5$ learners.}
\label{tab:NER-parameters}
\end{table*}

\begin{algorithm*}[h!]
\caption{Ensemble QA Prediction}
\label{alg:Ensemble QA Prediction}
\begin{algorithmic}[1]

\Require $k$ models $\{M_1, M_2, \ldots, M_k\}$, input question $q$
\Ensure Final answer $\hat{a}$ for the question $q$

\Statex
    \State $\mathcal{C} \gets \emptyset$ 

    \For{each model $M_i$ in $\{M_1, M_2, \ldots, M_k\}$}
        \State $\mathcal{A}_i \gets M_i(q)$ \Comment{Get candidate answers and their confidences}
        \For{each candidate $a_j$ in $\mathcal{A}_i$}
            \State $\mathcal{C} \gets \mathcal{C} \cup \{a_j\}$
        \EndFor
    \EndFor

    \State $\mathcal{C} \gets \left\{ c \mid \sum_{i=1}^{k} \mathbf{1}_{c \in \mathcal{A}_i} = k \right\} $ \Comment{Keep the candidates that appear in all models}

    \For{each candidate $c$ in $\mathcal{C}$}
        \State $\text{conf}_c \gets \frac{1}{k} \sum_{i=1}^k \text{confidence}_{M_i}(c)$ \Comment{Compute average confidence}
    \EndFor

    \State $\hat{a} \gets \arg\max_{c \in \mathcal{C}} \text{conf}_c$ \Comment{Select candidate with highest confidence}

    \State \Return $\hat{a}$

\end{algorithmic}
\end{algorithm*}

\label{sec:appendix}

\end{document}